\documentclass[letterpaper]{article} 
\usepackage{aaai2026}  
\usepackage{times}  
\usepackage{helvet}  
\usepackage{courier}  
\usepackage[hyphens]{url}  
\usepackage{graphicx} 
\usepackage{natbib}  
\usepackage{caption} 
\usepackage{algorithm}
\usepackage{algpseudocode}
\usepackage{algorithmicx}
\usepackage{amsmath}
\usepackage{amssymb}
\usepackage{subfig}

\usepackage{multirow}
\usepackage{newfloat}
\usepackage{listings}

\DeclareCaptionStyle{ruled}{labelfont=normalfont,labelsep=colon,strut=off} 
\floatstyle{ruled}
\newfloat{listing}{tb}{lst}{}
\floatname{listing}{Listing}

\title{Multi-Agent Trust Region Policy Optimisation: A Joint Constraint Approach}

\author{%
  Chak Lam Shek$^{1}$, Guangyao Shi$^{2}$,  Pratap Tokekar$^{1}$ 
}

\affiliations{
    $^{1}$University of Maryland \\ 
     $^{2}$ University of Southern California \\  
   
    \{cshek1, tokekar\}@umd.edu, shig@usc.edu
    
}

\begin{document}
\maketitle
\begin{abstract}
Multi-agent reinforcement learning (MARL) requires coordinated and stable policy updates among interacting agents. Heterogeneous-Agent Trust Region Policy Optimization (HATRPO) enforces per-agent trust region constraints using Kullback–Leibler (KL) divergence to stabilize training. However, assigning each agent the same KL threshold can lead to slow and locally optimal updates, especially in heterogeneous settings. To address this limitation, we propose two approaches for allocating the KL divergence threshold across agents: HATRPO-W, a Karush–Kuhn–Tucker-based (KKT-based) method that optimizes threshold assignment under global KL constraints, and HATRPO-G, a greedy algorithm that prioritizes agents based on improvement-to-divergence ratio. By connecting sequential policy optimization with constrained threshold scheduling, our approach enables more flexible and effective learning in heterogeneous-agent settings. Experimental results demonstrate that our methods significantly boost the performance of HATRPO, achieving faster convergence and higher final rewards across diverse MARL benchmarks. Specifically, HATRPO-W and HATRPO-G achieve comparable improvements in final performance, each exceeding 22.5\%. Notably, HATRPO-W also demonstrates more stable learning dynamics, as reflected by its lower variance.
\end{abstract}

\section{Introduction}

Multi-Agent Reinforcement Learning (MARL) is a critical area of artificial intelligence whose goal is to enable multiple agents to make decisions and coordinate within shared environments~\cite{canese2021multi}. Its versatility has driven impactful applications in domains such as robotics~\cite{gu2023safe, yu2021optimizing}, autonomous driving~\cite{wu2023intent, bhalla2020deep}, and smart grid management~\cite{park2022multi, roesch2020smart}, where effective coordination and adaptability are crucial. More recently, MARL techniques have been leveraged to enhance Large Language Models, exemplified by methods such as Group Relative Policy Optimization (GRPO) for coordinating multiple reasoning agents during text generation~\cite{shao2024deepseekmath}. These advancements underscore the importance of developing efficient and scalable MARL algorithms to propel progress in both traditional control systems and emerging foundation models.

In single-agent reinforcement learning, foundational methods such as Deep Q-Learning~\cite{mnih2015human} have laid the groundwork for effective policy learning in high-dimensional environments. Trust-region-based methods such as Trust Region Policy Optimization (TRPO)~\cite{schulman2015trust} and Proximal Policy Optimization (PPO)~\cite{schulman2017proximal} have demonstrated strong empirical performance, particularly in continuous control tasks. These methods have become the canonical approaches for RL. Inspired by these successes, researchers have extended these algorithms to the multi-agent setting, leading to variants such as Independent PPO (IPPO)~\cite{de2020independent} and Multi-Agent PPO (MAPPO)~\cite{yu2022surprising}. Although effective, these methods often suffer from gradient interference when multiple agents update their policies simultaneously, which can result in convergence to suboptimal local minima. 

To address this issue of gradient interference between agents, Heterogeneous-Agent Trust Region Policy Optimization  (HATRPO)~\cite{kuba2021trust} introduces a sequential policy update strategy, allowing agents to update one at a time. This reduces gradient conflicts and improves learning stability in heterogeneous multi-agent environments.


Despite the advancements brought by HATRPO, it assumes a uniform KL divergence threshold for all agents, which may not be optimal in heterogeneous settings where agents have varying capacities for improvement. This uniform allocation can limit overall system performance, as some agents might benefit from larger policy updates. To address this limitation, we propose two approaches for allocating the KL divergence threshold among agents: HATRPO-G, a greedy algorithm that prioritizes agents based on their improvement-to-divergence ratio, and HATRPO-W, a method based on Karush-Kuhn-Tucker (KKT) conditions to optimize threshold assignment under KL constraints. These methods aim to enhance the flexibility and effectiveness of policy updates in heterogeneous-agent environments. 

Our study investigates whether such adaptive KL divergence threshold allocation can lead to improved performance, more structured learning dynamics, and faster convergence, all while maintaining the same total KL divergence budget. Through this lens, we examine whether the learned threshold allocation reflects meaningful distinctions in agent advantage and whether it can overcome the limitations of fixed-threshold designs in coordinated multi-agent learning.

This paper makes the following key contributions:

\begin{itemize}
    \item We show that uniform KL thresholding across agents is suboptimal in sequential MARL, and propose a Joint KL constraint optimization problem to enable threshold allocation that improves overall coordination.
    
    \item We propose two methods for adaptive threshold allocation: a greedy method based on the improvement-to-divergence ratio, and a KKT-based optimization method that allocates the KL threshold under global constraints.

    \item Through extensive experiments on MARL benchmarks, we show that our adaptive KL allocation methods improve HATRPO's performance over baselines. Our results highlight that prioritizing agents based on their improvement potential leads to faster convergence, more effective policy updates, and better utilization of the KL budget. Specifically, HATRPO-W and HATRPO-G achieve comparable improvements in final performance, each exceeding 22.5\%. Notably, HATRPO-W also demonstrates more stable learning dynamics, as reflected by its lower variance.

\end{itemize}

\section{Related Work}
Early MARL methods predominantly assumed \emph{homogeneous} agents sharing policy parameters. Independent Q-Learning (IQL)~\cite{de2020independent}, Weighted QMIX (WQMIX)~\cite{rashid2020weighted}, QTRAN~\cite{son2019qtran, son2020qtran++}, and MADDPG~\cite{lowe2017multi} achieve strong performance in discrete cooperative tasks (e.g., SMAC) \cite{rashid2020monotonic}. While shared-policy approaches simplify cooperation, they inhibit expressive behavior when agents play \emph{heterogeneous} roles. The Heterogeneous-Agent Reinforcement Learning (HARL) framework addresses this by assigning each agent its own policy network, combined with centralized critics, sequential updates, and trust-region guarantees \cite{zhong2024heterogeneous, kuba2021trust}. In HARL, HATRPO (Heterogeneous-Agent TRPO) and HAPPO (Heterogeneous-Agent PPO) are two representative algorithms that are tractable and ensure monotonic improvement and joint-return guarantees \cite{kuba2021trust, zhong2024heterogeneous}. The theoretical foundation is strengthened by Heterogeneous-Agent Mirror Learning (HAML) \cite{kuba2022heterogeneous}, which guarantees both monotonic performance improvement and convergence to Nash equilibria; it also spawns additional agents like HAA2C, HADDPG, and HATD3 \cite{zhong2024heterogeneous, liu2023maximum}. Zhong et al.~\cite{zhong2024heterogeneous} propose the Multi-Agent Transformer (MAT), framing MARL as a sequential decision process in which a Transformer generates agent actions one-by-one to enhance coordination. Building on trust-region methods, A2PO~\cite{wang2023order} introduces an agent-by-agent policy optimization scheme that highlights the role of update ordering in achieving superior performance. To promote exploration in heterogeneous environments, MADPO~\cite{dou2024measuring} maximizes mutual policy divergence by explicitly distinguishing between intra- and inter-agent differences. FP3O~\cite{feng2023fp3o} extends PPO to support versatile parameter-sharing structures while preserving monotonic joint policy improvement guarantees. Complementary to these approaches, an optimism-driven gradient method~\cite{zhao2023optimistic} has been developed to improve sample efficiency in cooperative tasks. In parallel, TSPPO~\cite{tao2025tsppo} leverages Transformer-based sequential modeling within a PPO framework to improve agent coordination. Further enhancing Transformer integration, the Sequence Value Decomposition Transformer~\cite{zhao2025sequence} aligns agent-specific returns with cooperative objectives. Additionally, Liu et al.~\cite{liu2024jointppo} explore the role of joint optimization in PPO-based multi-agent learning. Finally, Zhang et al.~\cite{zhang2024sequential} propose a Stackelberg Decision Transformer to address asynchronous action dependencies through a sequential coordination framework.

In contrast, our work introduces a KL-constrained scheduling mechanism for HATRPO that adapts each agent's policy update magnitude based on its improvement potential. By formulating the update process as a joint KL optimization problem and introducing both greedy and KKT-based threshold allocation strategies, our approach addresses the limitations of uniform KL assignments and enhances both convergence speed and final performance in sequential MARL.


\section{PRELIMINARIES}
\label{PRELIMINARIES}
We consider a \textbf{Markov game}, defined by the tuple \( \mathcal{M} = \langle \mathcal{N}, \mathcal{S}, \mathcal{A}, \mathcal{P}, r, \gamma \rangle \), where \( \mathcal{N} = \{1, \dots, n\} \) is the set of \( n \) agents, \( \mathcal{S} \) is the finite state space, and \( \mathcal{A} = \prod_{i=1}^n \mathcal{A}_i \) is the joint action space with \( \mathcal{A}_i \) denoting the action space of agent \( i \). The transition dynamics are governed by the probability function \( \mathcal{P}(s' \mid s, \mathbf{a}) \), which specifies the likelihood of transitioning to state \( s' \) from state \( s \) under a joint action \( \mathbf{a} \in \mathcal{A} \). The reward function \( r(s, \mathbf{a}) \) returns a scalar reward for a given state and joint action, and \( \gamma \in [0, 1) \) is the discount factor. At each time step \( t \), the agents occupy a state \( s_t \in \mathcal{S} \) and each agent \( i \in \mathcal{N} \) selects an action \( a_t^i \in \mathcal{A}_i \) according to its policy \( \pi_i(\cdot \mid s_t) \). These actions form the joint action \( \mathbf{a}_t = (a_t^1, \dots, a_t^n) \in \mathcal{A} \), drawn from the joint policy \( \pi(\cdot \mid s_t) = \prod_{i=1}^n \pi_i(\cdot \mid s_t) \). The agents then receive a joint reward \( r_t = r(s_t, \mathbf{a}_t) \) and transition to the next state \( s_{t+1} \), sampled according to \( \mathcal{P}(s_{t+1} \mid s_t, \mathbf{a}_t) \).

The joint policy \( \pi \),  transition function \( \mathcal{P} \), and initial state distribution \( \rho_0 \) together induce a sequence of marginal state distributions \( \rho_t^\pi \) at each timestep \( t \). The overall discounted state visitation distribution is defined as:
$
\rho^\pi = \sum_{t=0}^\infty \gamma^t \rho_t^\pi.
$

The value of a state under policy \( \pi \) is given by the state value function:
$
V^\pi(s) = \mathbb{E} \left[ \sum_{t=0}^\infty \gamma^t r_t \,\middle|\, s_0 = s \right],
$ 
and the expected return for executing a joint action \( \mathbf{a} \) from state \( s \) is captured by 
$
Q^\pi(s, \mathbf{a}) = \mathbb{E} \left[ \sum_{t=0}^\infty \gamma^t r_t \,\middle|\, s_0 = s, \mathbf{a}_0 = \mathbf{a} \right].
$

The corresponding advantage function, which quantifies the benefit of taking joint action \( \mathbf{a} \) over following the policy \( \pi \), is defined as:
$
A^\pi(s, \mathbf{a}) = Q^\pi(s, \mathbf{a}) - V^\pi(s).
$

In this setting, all agents share the same reward function. The goal is to maximize the expected total reward:  
$
J(\pi) = \mathbb{E} \left[ \sum_{t=0}^\infty \gamma^t r_t \right],  
$  
where \( s_0, s_1, \dots \sim \rho^\pi \) and \( \mathbf{a}_0, \mathbf{a}_1, \dots \sim \pi \).  

We also investigate the contribution to performance from different subsets of agents. Let \( i_{1:m} \) denote an ordered subset \( \{i_1, \dots, i_m\} \) of \( \mathcal{N} \), and \( -i_{1:m} \) refer to its complement. The \( k \)-th agent in the subset is written as \( i_k \).  

The \textbf{multi-agent state-action value function} is defined as:  
$
Q_{i_{1:m}}^\pi(s, \mathbf{a}) = \mathbb{E}_{\mathbf{a}_{-i_{1:m}} \sim \pi_{-i_{1:m}}} \left[ Q^\pi(s, \mathbf{a}_{i_{1:m}}, \mathbf{a}_{-i_{1:m}}) \right].  
$

For disjoint sets \( j_{1:k} \) and \( i_{1:m} \), the \textbf{multi-agent advantage function} is defined as:  
\begin{equation*}
   \resizebox{\columnwidth}{!}{$
A_{i_{1:m}}^\pi(s, \mathbf{a}_{j_{1:k}}, \mathbf{a}_{i_{1:m}}) = Q_{j_{1:k}, i_{1:m}}^\pi(s, \mathbf{a}_{j_{1:k}}, \mathbf{a}_{i_{1:m}}) - Q_{j_{1:k}}^\pi(s, \mathbf{a}_{j_{1:k}}). $
} 
\end{equation*}

Here, the joint policies \( \pi = (\pi_1, \dots, \pi_n) \) and \( \bar{\pi} = (\bar{\pi}_1, \dots, \bar{\pi}_n) \) represent the “current” and “new” joint policies, respectively, as agents update their strategies.

\subsection{Heterogeneous-Agent TRPO Algorithm}

The \textit{Heterogeneous Agents Trust Region Proximal Optimization} (HATRPO) algorithm~\cite{zhong2024heterogeneous} extends Trust Region Policy Optimization (TRPO) to heterogeneous multi-agent systems. Unlike homogeneous-agent methods that assume shared observation and action spaces or identical policy architectures, HATRPO accommodates agents with distinct modalities and parameterizations.

A key feature of HATRPO is that it preserves the monotonic improvement guarantee at the \emph{joint policy level}, meaning that sequential updates of individual agents still lead to consistent improvement of the overall system performance. Each agent solves a constrained optimization problem to improve its local policy while ensuring bounded divergence from the previous policy.

The policy update for agent \( i_m \in i_{1:m} \) at iteration \( k+1 \) is given by the following constrained optimization:

\begin{equation}
\resizebox{\columnwidth}{!}{$
\begin{aligned}
\pi_{k+1}^{i_m} = \arg\max_{\pi^{i_m}} \;
& \mathbb{E}_{s \sim \rho_\pi,\; a^{i_{1:m-1}} \sim \pi^{i_{1:m-1}},\; a^{i_m} \sim \pi^{i_m}} 
\big[ A_{\pi}^{i_m}(s, \mathbf{a}^{i_{1:m-1}}, a^{i_m}) \big] \\
\text{s.t.} \;
& \mathbb{E}_{s \sim \rho_\pi} \big[ D_{\mathrm{KL}}\big( \pi_k^{i_m}(\cdot \mid s) \,\|\, \pi^{i_m}(\cdot \mid s) \big) \big] \leq \delta
\end{aligned}
$}
\label{eq:hatrpo-update}
\end{equation}

The surrogate objective used in this optimization is defined as:
\begin{equation}
\resizebox{\columnwidth}{!}{$
L_{\pi}^{i_{1:m}}(\bar{\pi}^{i_{1:m-1}}, \hat{\pi}^{i_m})
\triangleq 
\mathbb{E}_{s \sim \rho_\pi,\; a^{i_{1:m-1}} \sim \bar{\pi}^{i_{1:m-1}},\; a^{i_m} \sim \hat{\pi}^{i_m}}
\big[ A_{\pi}^{i_m}(s, \mathbf{a}^{i_{1:m-1}}, a^{i_m}) \big]
$}
\end{equation}

By leveraging centralized training with decentralized execution, HATRPO facilitates efficient, stable learning in multi-agent systems composed of heterogeneous agents while maintaining strong performance guarantees.

\section{Jointly Constrained HATRPO}

While HATRPO offers theoretical guarantees by enforcing trust region constraints for stable policy updates, it applies a uniform KL divergence threshold to each agent individually:
\begin{equation}
D_{\mathrm{KL}}^{\max}(\pi_i \,\|\, \bar{\pi}_i) \leq \delta, \quad \forall i = 1, \dots, m
\end{equation}
A small KL divergence threshold is often preferred in trust-region methods because it provides a tighter lower bound on policy performance improvement. By limiting how far a new policy can deviate from the old one, small \( \delta \) values ensure stable and theoretically grounded learning steps.

However, in HATRPO, assigning the same small \( \delta \) to every agent can inadvertently slow down the overall learning process. Specifically, agents with low advantage values or those operating in flat regions of the optimization landscape may receive updates that are minimal or unproductive. This restricts their ability to escape poor local optima and contributes to inefficient use of the policy update threshold.

\begin{figure}[ht]
    \centering
   \includegraphics[width=0.75\linewidth]{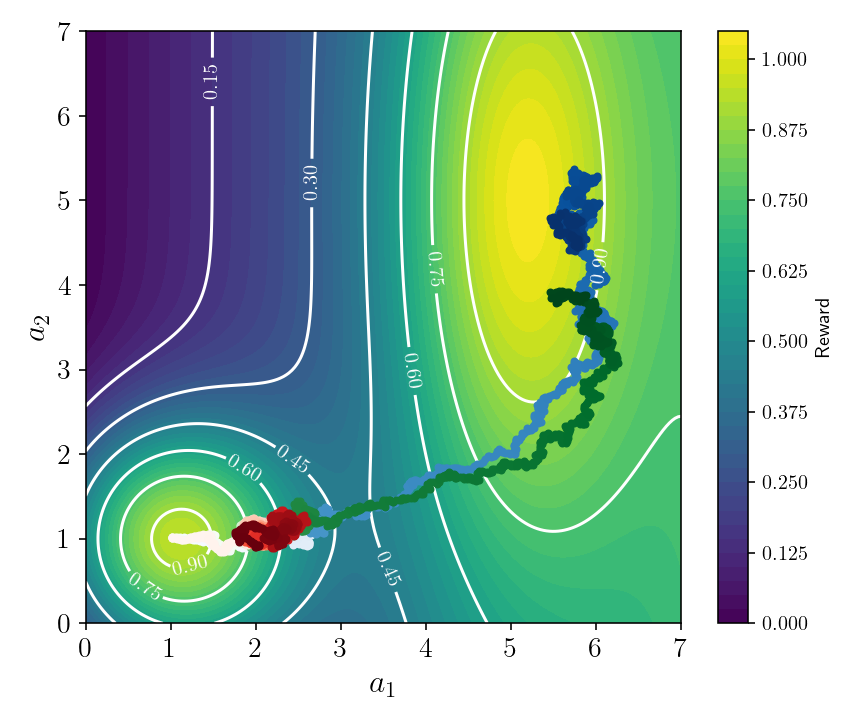}
   \vspace{-7pt}
\caption{Heatmap of the reward function
        $R(a_1, a_2) = 10\,\mathcal{N}(5,5;\,1,3) + 0.1 a_1 + 5.3\, \mathcal{N}(1,1;\,1,1)$,
        where \( \mathcal{N}(\mu_1, \mu_2;\, \sigma_1, \sigma_2) \) denotes a 2D Gaussian with mean \( (\mu_1, \mu_2) \) and standard deviation \( (\sigma_1, \sigma_2) \). The plot shows the joint reward surface and the action trajectories of two players optimizing their actions \( a_1 \) and \( a_2 \). Color gradients indicate optimization progress: \textcolor{red}{red} for HATRPO, \textcolor{blue}{blue} for HATRPO-G, and \textcolor{green!60!black}{green} for HATRPO-W.
    }
    \vspace{-10 pt}
    \label{fig:Guassian_heatmap}
\end{figure}

This limitation is clearly illustrated in Figure~\ref{fig:Guassian_heatmap}, which depicts a two-player differential matrix game where the reward function consists of a combination of two Gaussian modes: one corresponding to a local optimum and the other to a global optimum with broader spread in the \( a_2 \) (vertical) direction. The policies are initialized as Gaussians with means at \( (a_1, a_2) = (1,1) \), placing them near the local optimum. Under HATRPO, both players are constrained by the same KL threshold, which prevents sufficient exploration, especially by player 1, and leads the joint policy to remain trapped near the local optimum.

In contrast, HATRPO-G and HATRPO-W dynamically adjust the KL threshold allocation, allowing player 1 to receive a larger update allowance. Since the global optimum has low curvature in the \( a_2 \) direction, increasing the KL threshold for player 1 allows the joint policy to escape the local optimum and converge to the global maximum. 

Moreover, since each agent is constrained independently, the uniform allocation does not account for varying contributions to global performance. As a result, the overall system improvement may be suboptimal, especially in heterogeneous settings where agents differ in learning potential or strategic significance.

This constraint also directly affects the speed of convergence. To illustrate, consider a multi-agent matrix game where each of the \( N \) agents simultaneously selects an action \( a_i \in \{0,1\} \). The joint reward is defined as follows:
\begin{equation}
\resizebox{0.45\textwidth}{!}{$
R(\mathbf{a}) =
\begin{cases}
1.5 & \text{if } a_i = 1 \text{ for all } i,\\
1   & \text{if } \exists j \text{ s.t. } a_j = 0 \text{ and } a_k = 0 \; \forall k > j, \\
0   & \text{otherwise.}
\end{cases}
$}
\end{equation}

In this setting, the optimal reward (1.5) is only achieved when all agents simultaneously select 1. However, the reward function is inherently non-symmetric. Low-index agents are more likely to receive positive advantage estimates because a later agent can only obtain reward if all earlier agents choose action 1. As a result, during the early learning phase, low-index agents tend to accumulate positive gradients and request substantial policy updates, while high-index agents experience little or no improvement signal and remain relatively static.

\begin{figure}[ht]
    \centering
    \includegraphics[width=0.8\linewidth]{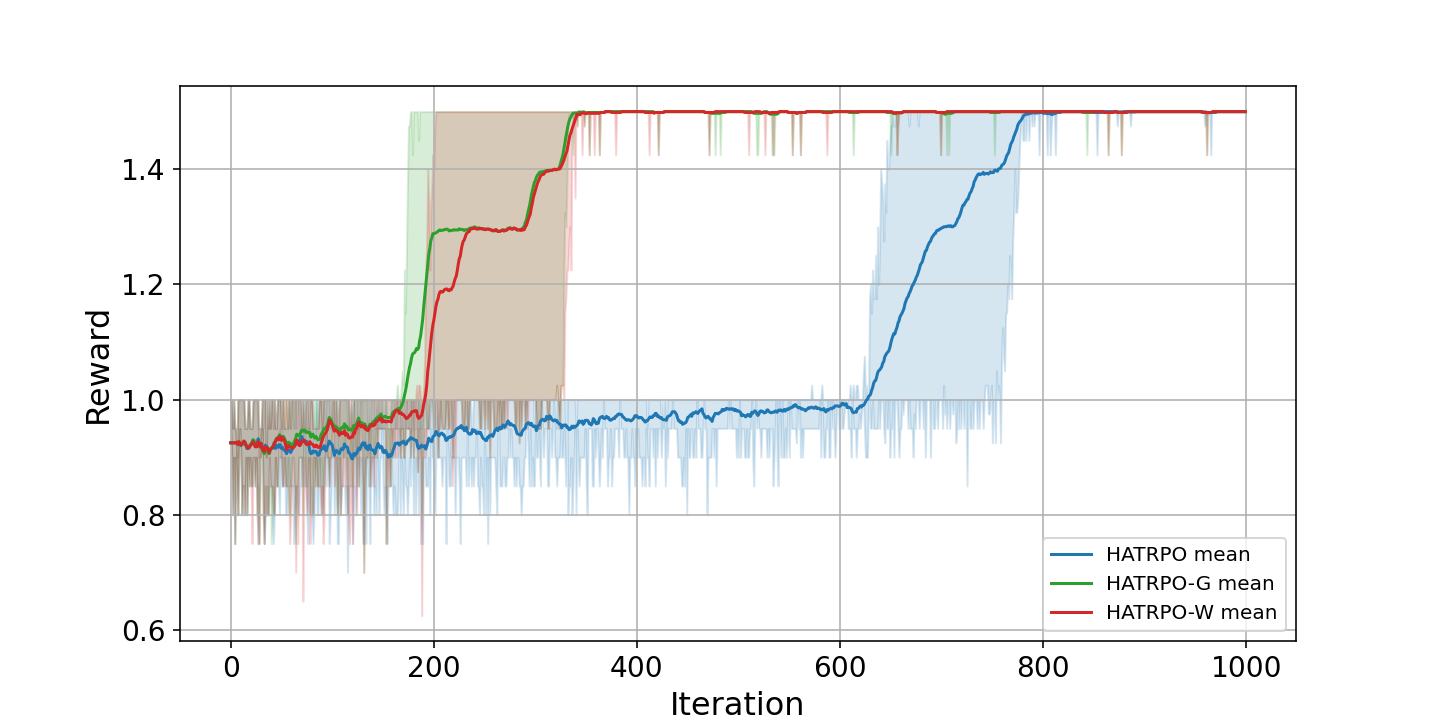}
    \caption{Average reward as a function of training iterations in the 4-agent matrix game.}
    \vspace{-7pt}
    \label{fig:Matrix_reward}
\end{figure}

This imbalance leads to a mismatch between policy improvement demand and update capacity when a fixed KL threshold is distributed uniformly. Consequently, the overall learning progress is bottlenecked by high-index agents that adapt slowly, which hinders the system's ability to coordinate toward the optimal joint policy. This phenomenon is reflected in the learning curve behavior shown in Figure~\ref{fig:Matrix_reward}.
These observations motivate the need for a more flexible allocation of policy update thresholds, which can accelerate convergence and enable coordinated breakthroughs in complex multi-agent interactions.
\subsection{Problem Formulation}
To address this, we reformulate the constraint structure while retaining the same objective. We continue to maximize the total advantage across all agents:
\begin{equation}
\max_{\pi_1, \dots, \pi_m} \quad \sum_{i=1}^{m} \mathcal{L}_i(\bar{\pi}_{1:i-1}, \pi_i)
\end{equation}
But instead of enforcing per-agent KL constraints, we propose a global trust region:
\begin{equation}
\sum_{i=1}^{m} D_{\mathrm{KL}}^{\max}(\pi_i \,\|\, \bar{\pi}_i) \leq \delta_{\text{total}}
\end{equation}
This formulation allows agents to share a common KL threshold and redistribute improvement potential more effectively---allocating more threshold to high-impact agents and reducing waste. Agents with larger advantage functions and smaller policy changes are more desirable for updates. Additionally, since KL divergence is not symmetric and policy order affects downstream value changes, update order matters~\cite{wen2022game}. This prioritization enables better group-level coordination and maximizes collective reward.
\begin{algorithm}[H]
\caption{Modified HATRPO with Hard KL Constraint and KL Divergence Threshold Allocation}
\label{alg:hatrpo}
\begin{algorithmic}[1]  
\Require Initial joint policy \( \pi_0 = (\pi_0^1, \dots, \pi_0^m) \)
\For{$k = 0, 1, 2, \dots$}
    \State Compute the joint advantage function \( A^{\pi_k}(s, a) \) for all state-action pairs
    \State  Compute  \(Order( i_{1:m}\)), \(\{ \delta_1, ..., \delta_m \}\) using  KL allocation algorithm \ref{alg:HATRPO-G} or \ref{alg:HATRPO-W}
    \State \textit{// Solve the following constrained sequential optimization problem \ref{eq:hatrpo-update} with $\delta_i$}

\EndFor
\end{algorithmic}
\end{algorithm}
\vspace{-15 pt}

\begin{algorithm}[H]
\caption{Greedy Agent Ordering via Score-Based Ranking}
\label{alg:HATRPO-G}
\begin{algorithmic}[1]
\Require Fixed priors \( \bar{\pi} \), total KL threshold \( \delta_{\text{total}} \), loss functions \( \mathcal{L}_i \), KL bounds \( D^{\max}_{\text{KL}}(\pi_i \| \bar{\pi}_i) \), constant \( \epsilon > 0 \)
\State Initialize ordered list \( \mathcal{O} \gets [] \), remaining agents \( \mathcal{R} \gets \mathcal{A} \)
\While{\( \mathcal{R} \neq \emptyset \) and $\delta_{\text{total}} > 0$}
    \ForAll{\( i \in \mathcal{R} \)}
        \State Solve
        \Statex \hspace{\algorithmicindent}
        \resizebox{0.7\columnwidth}{!}{$
        \begin{aligned}
            \max_{\pi_i} \; & \mathcal{L}_i(\bar{\pi}_{1:i-1}, \pi_i) \\
            \text{s.t.} \quad & \mathbb{E}_{s \sim d^{\pi_k}} \left[ D_{\mathrm{KL}}\left(\pi_i(\cdot|s) \,\|\, \pi_k^i(\cdot|s)\right) \right] \leq \delta_{total}
        \end{aligned}$
        }

       \State  Compute score:
        $\text{Score}_i \gets \frac{\mathcal{L}_i(\bar{\pi}_{1:i-1}, \pi_i)}{D^{\max}_{\text{KL}}(\pi_i \| \bar{\pi}_i) + \epsilon}$
        
    \EndFor
    \State Select agent \( i^* \gets \arg\max_{i \in \mathcal{R}} \text{Score}_i \)
    \State Append \( i^* \) to \( \mathcal{O} \), fix \( \pi_{i^*} \), remove \( i^* \) from \( \mathcal{R} \)
    \State Update $\delta_{total} \gets \delta_{total} - D^{\max}_{\text{KL}}(\pi_i \| \bar{\pi}_i)$
\EndWhile
\end{algorithmic}
\end{algorithm}
\vspace{-10pt}
To improve coordination efficiency in multi-agent policy optimization, we modify the original HATRPO algorithm by introducing adaptive, agent-specific KL divergence thresholds. Instead of applying a uniform trust region constraint across all agents, our approach dynamically allocates the KL budget based on each agent's estimated contribution to overall performance. As shown in \textbf{Algorithm~\ref{alg:hatrpo}}, we incorporate a KL allocation strategy---either a greedy score-based ranking or a KKT-based optimization---that determines both the per-agent KL bounds and their update order. This enables the algorithm to sequentially optimize each agent’s policy within a tailored trust region.


\subsection{HATRPO-Greedy}

We first propose an effective greedy algorithm, HATRPO-G to determine the order of policy updates. The idea is to prioritize agents whose updates provide the highest benefit-to-cost ratio, measured as the advantage gain divided by the KL divergence. At each step, we select the agent with the best ratio and fix its policy for subsequent evaluations. As described in \textbf{Algorithm~2}, the method constructs an ordered list of agents by selecting, at each iteration, the agent that maximizes the expected improvement per unit of KL cost. Specifically, each agent \( i \) is assigned a score given by
\[
\text{Score}_i = \frac{\mathcal{L}_i(\bar{\pi}_{1:i-1}, \pi_i)}{D^{\max}_{\mathrm{KL}}(\pi_i \| \bar{\pi}_i) + \epsilon},
\]
where \( \mathcal{L}_i \) represents the estimated local policy improvement and \( \epsilon > 0 \) is a small constant for numerical stability. The agent with the highest score is selected, added to the update order, and removed from the candidate pool. The process repeats until all agents are ranked. This greedy procedure introduces a soft prioritization mechanism that favors agents expected to yield higher utility per KL cost, resulting in a scalable approach for determining the agent update sequence without requiring global coordination or joint optimization.

\subsection{HATRPO-Weighted}
In the HATRPO-W, we adapt the classical water-filling strategy~\cite{4151217,2013ITWC...12.3637H} from communications to formulate KL divergence allocation as a constrained optimization problem, solved using KKT conditions. Just as the water-filling method distributes power across noisy channels to maximize capacity, we allocate a total KL divergence budget across agents to maximize overall policy improvement. Agents with higher expected gain per unit KL—analogous to high-SNR channels—are assigned larger updates, while those with lower impact may receive little or none. This induces a soft prioritization mechanism that allocates the trust region more efficiently among agents. The resulting KKT-based allocation can be solved numerically (e.g., via bisection over the Lagrange multiplier) and yields a principled, globally coordinated update scheme under a shared KL constraint.

\begin{algorithm}[H]
\caption{KL Allocation via KKT}
\label{alg:HATRPO-W}
\begin{algorithmic}[1]
\Require Fixed priors \( \bar{\pi} \), advantage estimates \( A_i \), total KL threshold \( \delta_{\text{total}} \), tolerance \( \varepsilon \)
\State Initialize \( \delta_i \gets 0 \) for all agents \( i \)
\State Compute effective utility: \( U_i \gets \mathbb{E}_{a \sim \bar{\pi}_i}[A_i(a)] \)
\State \(Order(i_{1:m}) \) = Sort agents in decreasing order of \( U_i \)
\State Initialize \( \lambda \leftarrow \text{large positive value} \)
\Repeat
    \State Compute tentative allocation: \resizebox{0.35\columnwidth}{!}{
$ \delta_i \leftarrow \max\left(0, \frac{U_i}{\lambda} - 1\right)$
}
    
    \State Evaluate total KL: \( \delta = \sum_i \delta_i \)
    \State Update \( \lambda \leftarrow \lambda \cdot (\delta / \delta_{\text{total}}) \)
\Until{\( |\delta - \delta_{\text{total}}| < \varepsilon \)}
\State \Return $Order(i_{1:m}), \{\delta_i, ..., \delta_m \}$ 
\end{algorithmic}
\end{algorithm}

We relax the total KL constraint using a Lagrange multiplier \( \lambda \geq 0 \), yielding the following Lagrangian objective:
\begin{equation}
\max_{\pi_1, \dots, \pi_m} \quad \sum_{i=1}^{m} \left[ \mathcal{L}_i(\bar{\pi}_{1:i-1}, \pi_i) - \lambda \cdot C_i \right] - \lambda \delta_{\text{total}},
\end{equation}
where \( C_i \) denotes the KL divergence between the updated policy \( \pi_i \) and its prior \( \bar{\pi}_i \).

To derive the optimal allocation, we first approximate each agent's local objective \( \mathcal{L}_i \) using a first-order expansion around the prior policy:
\begin{equation}
\mathcal{L}_i(\bar{\pi}_{1:i-1}, \pi_i) \approx \sum_{a} A_i(a) \cdot \frac{\pi_i(a)}{\bar{\pi}_i(a)},
\end{equation}
where \( A_i(a) \) is the advantage function under the prior \( \bar{\pi}_i \). We similarly approximate the KL divergence:
\begin{equation}
C_i \approx \sum_{a} \left( \frac{\pi_i(a)^2}{\bar{\pi}_i(a)} - 1 \right).
\end{equation}

Optimizing the Lagrangian with respect to \( \pi_i \), we obtain the condition:
\begin{equation}
\frac{\partial \mathcal{L}_i}{\partial C_i} = \lambda.
\end{equation}
Assuming a linear relationship between utility and KL, we find a closed-form for the KL budget:
\begin{equation}
C_i \approx \frac{\mathbb{E}_{a \sim \bar{\pi}_i}[A_i(a)]}{\lambda} - 1.
\end{equation}
To ensure non-negativity of KL, we apply a projection:
\begin{equation}
C_i = \max\left(0, \frac{\mathbb{E}_{a \sim \bar{\pi}_i}[A_i(a)]}{\lambda} - 1 \right).
\end{equation}

To satisfy the global trust region constraint, the total allocated KL divergence must equal \( \delta_{\text{total}} \):
\begin{equation}
\sum_{i=1}^{m} C_i = \delta_{\text{total}}.
\end{equation}
Substituting the expression for \( C_i \), we obtain:
\begin{equation}
\sum_{i=1}^{m} \max\left(0, \frac{\mathbb{E}_{a \sim \bar{\pi}_i}[A_i(a)]}{\lambda} - 1 \right) = \delta_{\text{total}}.
\end{equation}
This nonlinear equation implicitly defines the optimal \( \lambda \) and can be efficiently solved using numerical methods such as bisection. Once \( \lambda \) is determined, each agent’s KL allocation \( C_i \) can be computed accordingly, resulting in an adaptive and coordinated trust region allocation strategy.

In practice, we compute the optimal KL allocation by solving for the Lagrange multiplier \( \lambda \) using an iterative bisection-style update. As shown in \textbf{Algorithm \ref{alg:HATRPO-W}}, we first compute each agent’s utility score \( U_i = \mathbb{E}_{a \sim \bar{\pi}_i}[A_i(a)] \), which measures the expected improvement under the prior policy. We initialize \( \lambda \) to a small positive value and repeatedly compute the tentative KL allocation for each agent using the closed-form:
$
D^{\max}_{\mathrm{KL}}(i) = \max\left(0, \frac{U_i}{\lambda} - 1\right).
$
This expression reflects the intuition that agents with higher utility receive more KL budget, while those with lower utility may receive none. We then evaluate the total KL allocation \( \delta = \sum_i D^{\max}_{\mathrm{KL}}(i) \) and adjust \( \lambda \) multiplicatively to match the target total KL threshold \( \delta_{\text{total}} \). This process is repeated until convergence, defined by a small absolute difference \( |\delta - \delta_{\text{total}}| < \varepsilon \). This method provides a efficient solution to the KL allocation problem and scales easily to many agents.

\section{RESULTS}
\begin{figure*}[ht]
    \centering
    \subfloat[HATRPO]{
        \includegraphics[width=0.28\textwidth]{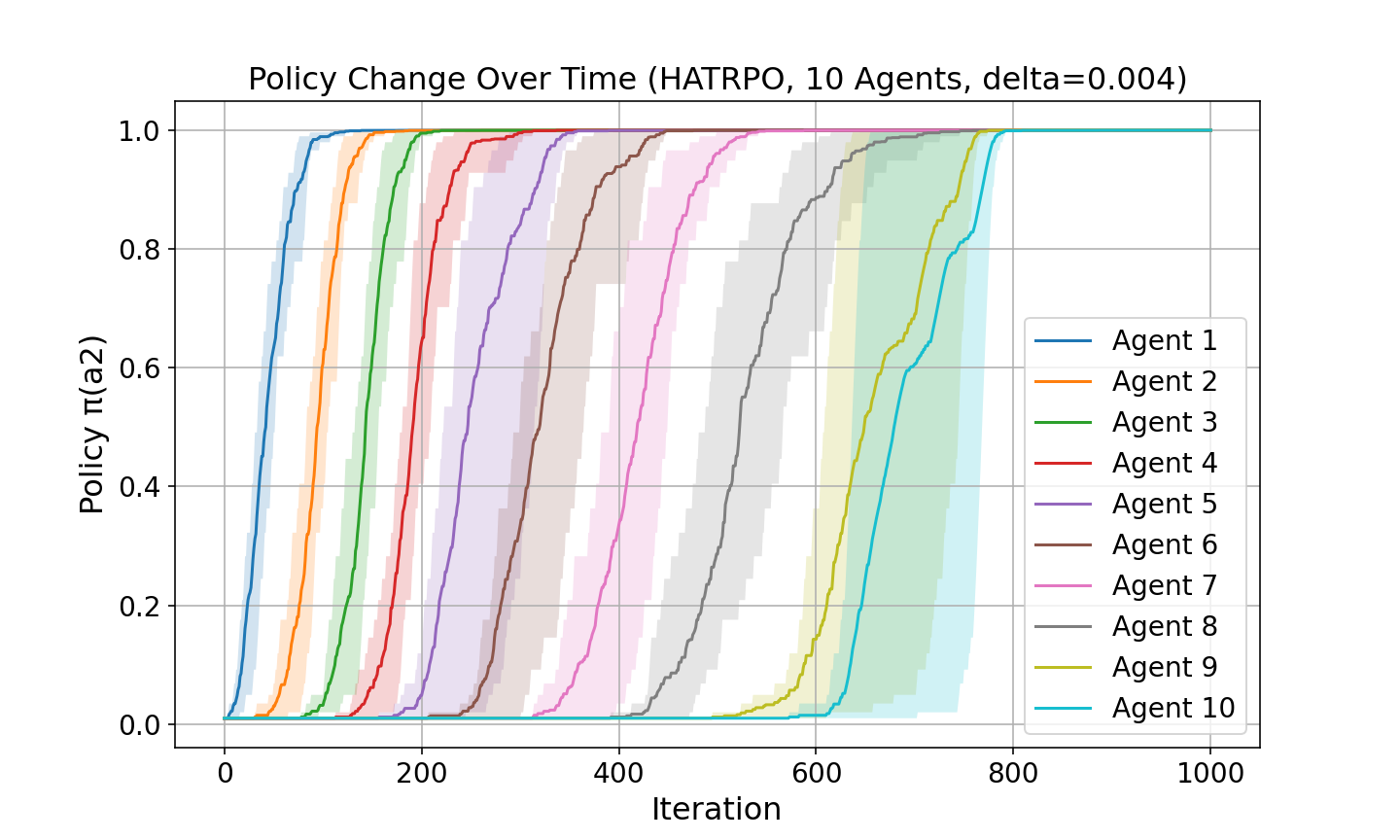}
        \label{fig:Matrix_HATRPO}
    }
    \subfloat[HATRPO-G]{
        \includegraphics[width=0.28\textwidth]{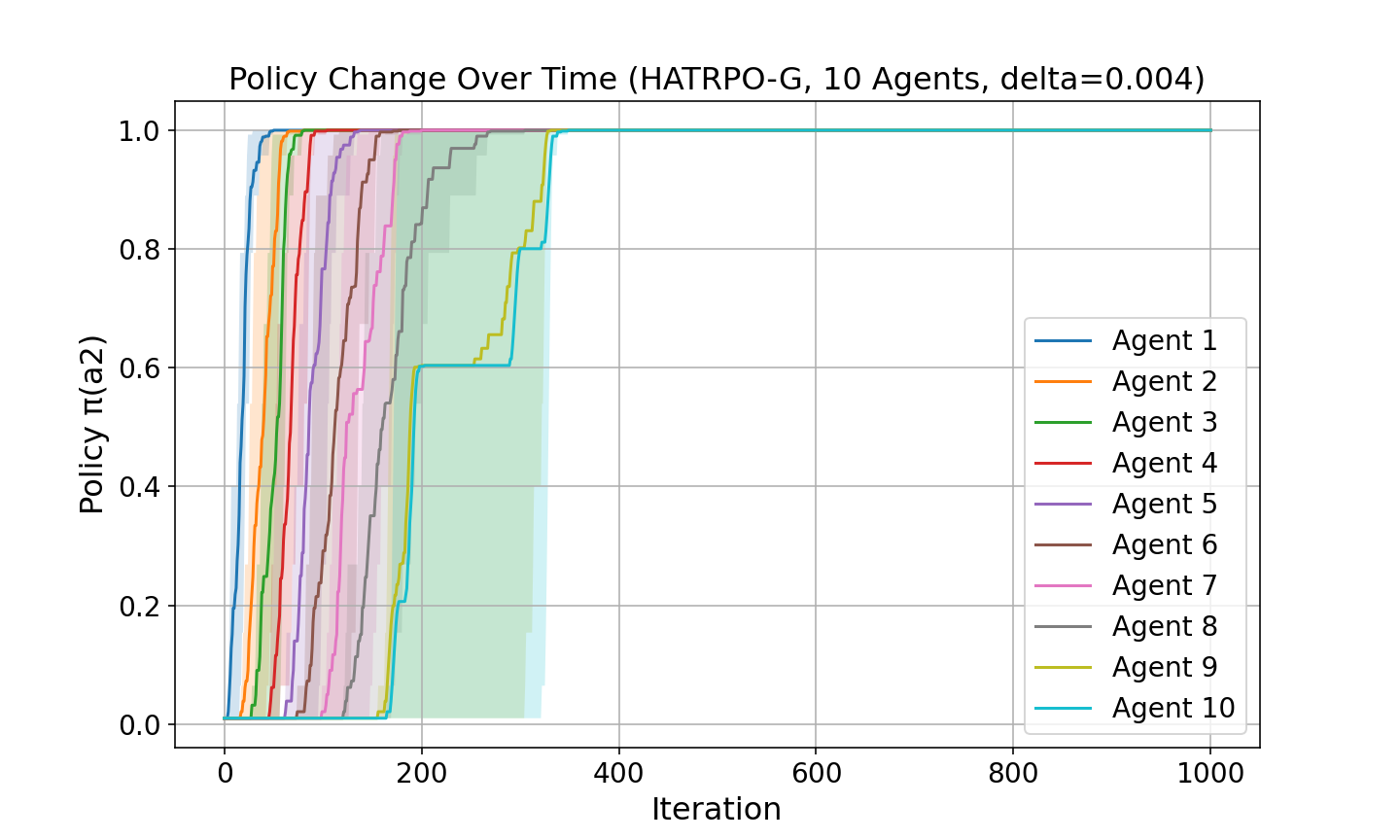}
        \label{fig:Matrix_HATRPO-G}
    }
    \subfloat[HATRPO-W]{
        \includegraphics[width=0.28\textwidth]{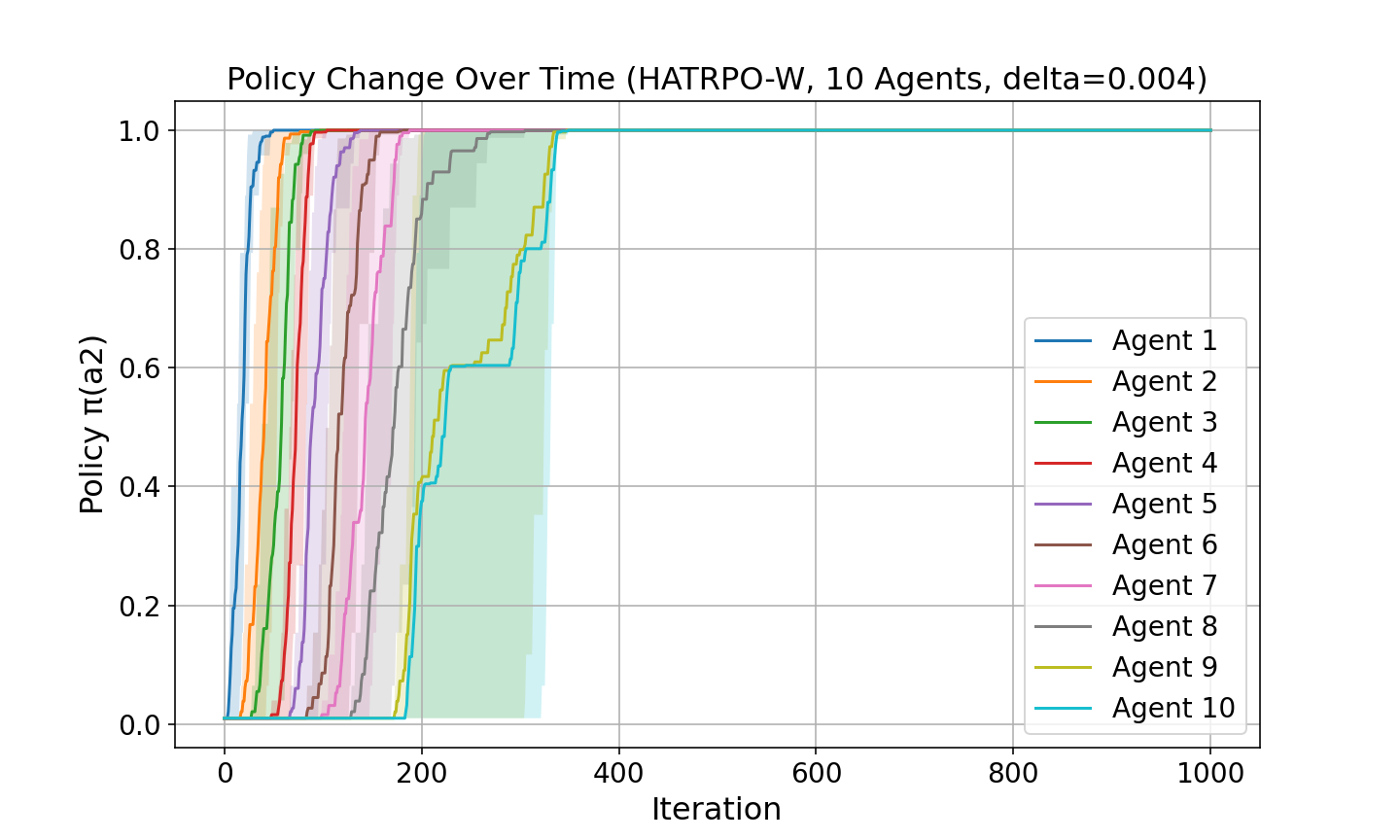}
        \label{fig:Matrix_HATRPO-W}
    }
    \caption{
        Policy trajectory evolution for all agents under different algorithms in the 10-agent matrix game.
    }
    \vspace{-14pt}
    \label{fig:Matrix_Policy}
\end{figure*}

\begin{figure*}[ht]
    \centering
    \subfloat[HATRPO]{
        \includegraphics[width=0.28\textwidth]{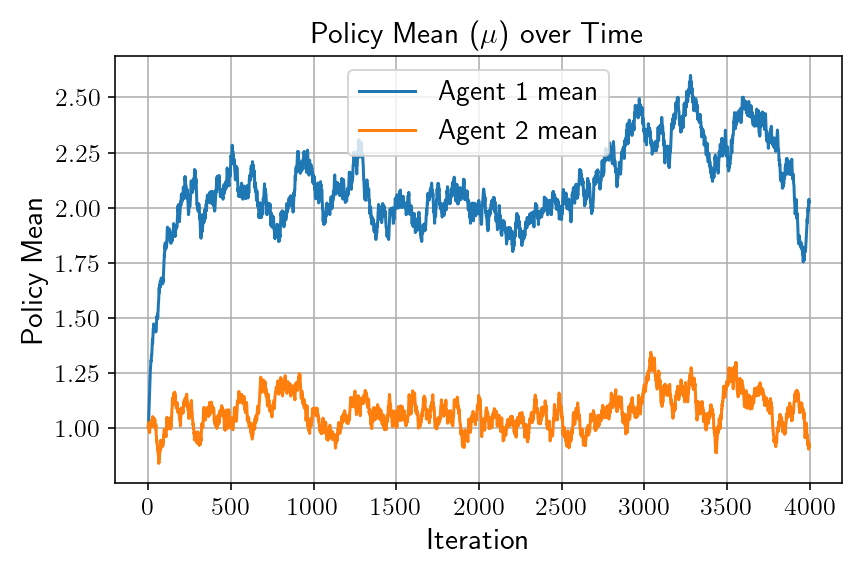}
        \label{fig:gaussion_HATRPO}
    }
    \subfloat[HATRPO-G]{
        \includegraphics[width=0.28\textwidth]{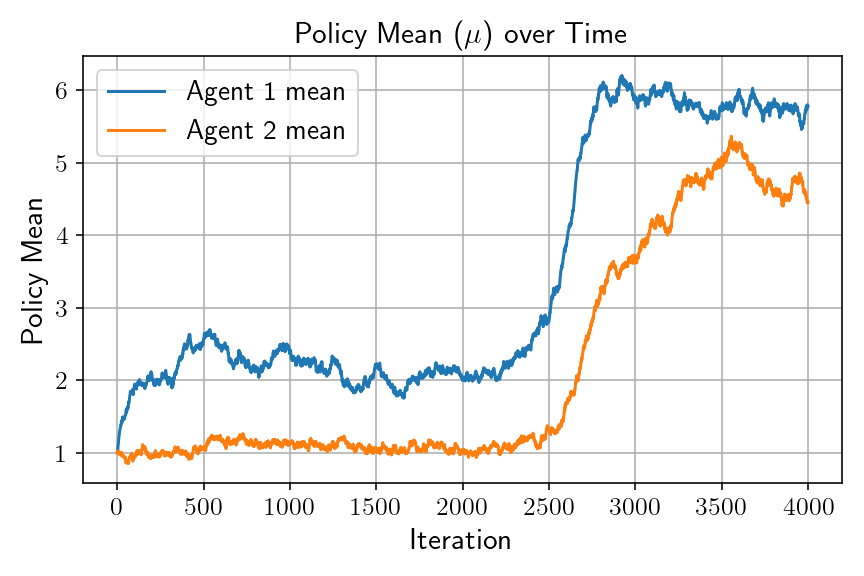}
        \label{fig:gaussion_HATRPO-G}
    }
    \subfloat[HATRPO-W]{
        \includegraphics[width=0.28\textwidth]{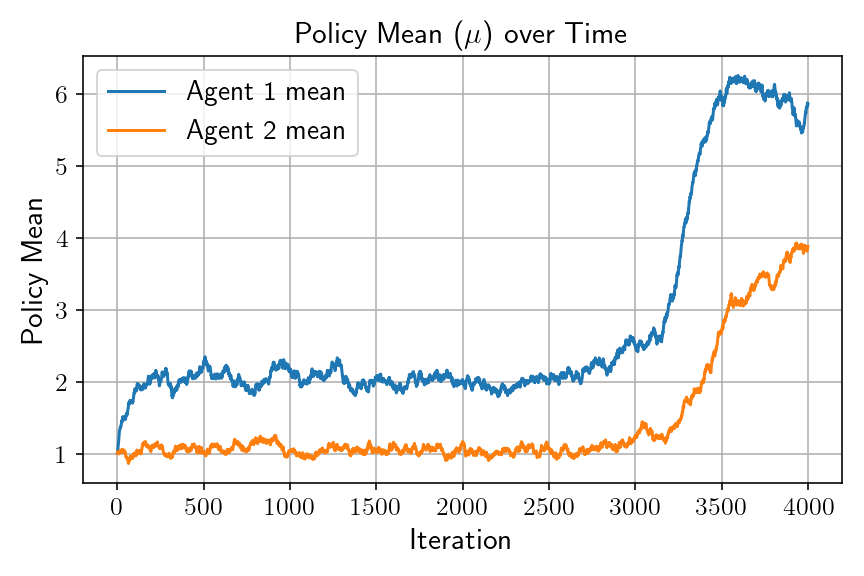}
        \label{fig:gaussion_HATRPO-W}
    }
    \caption{
        Policy trajectory evolution for all agents under different algorithms in the 2-player differential game.
    }
    \vspace{-13 pt}
    \label{fig:Gaussian_Policy}
\end{figure*}
 We evaluate our proposed methods on three MARL settings of increasing complexity. 
 1) \textbf{Matrix Game: Coordination Under Sparse Reward} to study how KL allocation affects convergence speed in the presence of non-uniform advantage distributions.
 2) \textbf{Differential Game: Escaping Local Optima} to test the ability of adaptive KL strategies to escape local optima.
 3) \textbf{Multi-Agent MuJoCo: Realistic Heterogeneous Agents} to assess the scalability and adaptability of our methods in realistic settings. 

\noindent \textbf{Baselines.} We compare our methods against several representative MARL algorithms: 1) \textbf{MADDPG}~\cite{lowe2017multi}; 2) \textbf{MAPPO}~\cite{yu2022surprising}; 3) \textbf{HAPPO}~\cite{kuba2021trust}; 
4) \textbf{HATRPO}~\cite{kuba2021trust}.

Implementation details, MARL benchmark descriptions, and hyperparameter settings are provided in Appendix A. We aim to answer the following research questions.


\subsubsection{(a) How does our algorithm compare to strong baselines under the same total KL divergence threshold?}
\begin{figure*}[ht]
    \centering
    \begin{tabular}{cccc}
        \subfloat[2x4-Agent Ant]{%
            \includegraphics[width=0.22\textwidth]{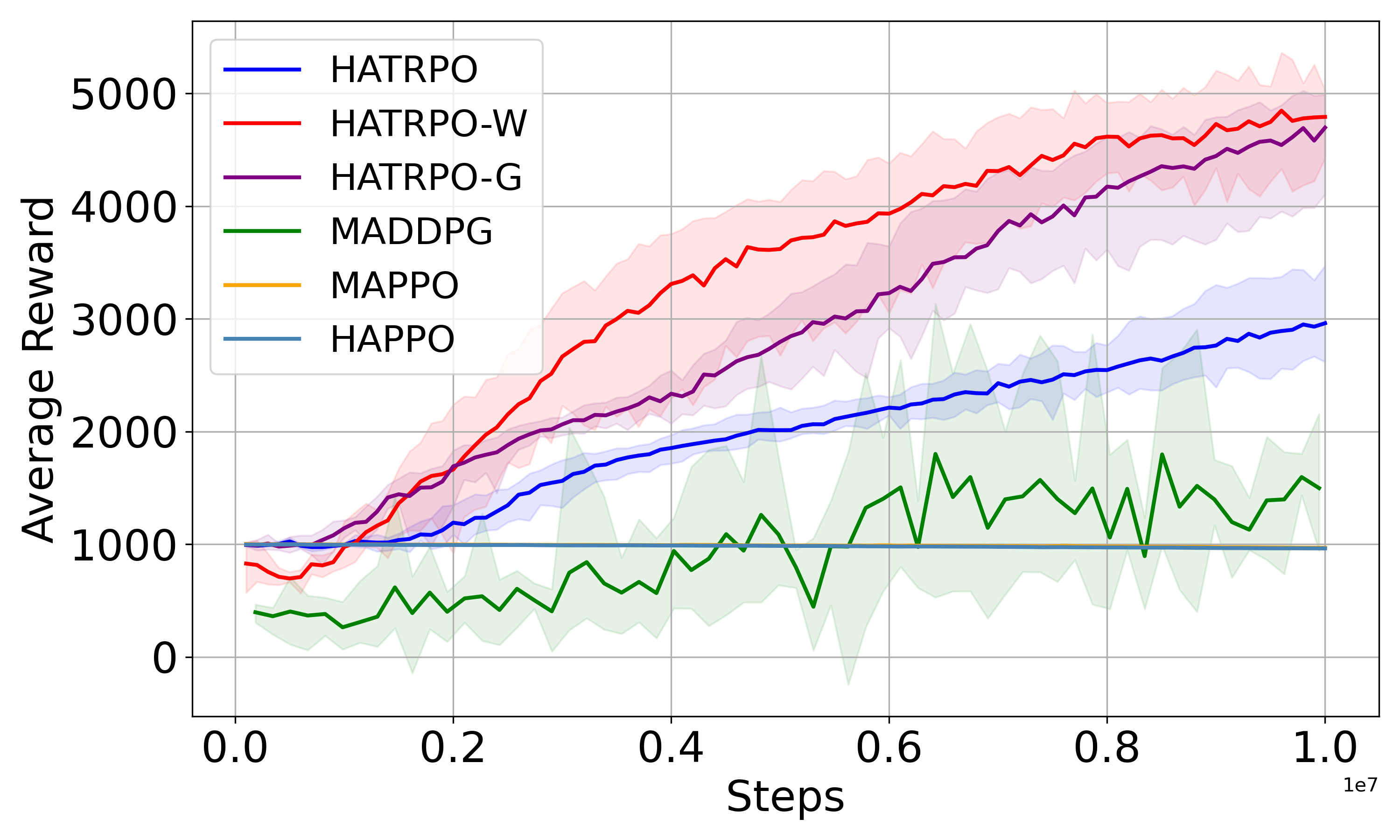}
            \label{fig:fig1}
        } &
        \subfloat[4x2-Agent Ant]{%
            \includegraphics[width=0.22\textwidth]{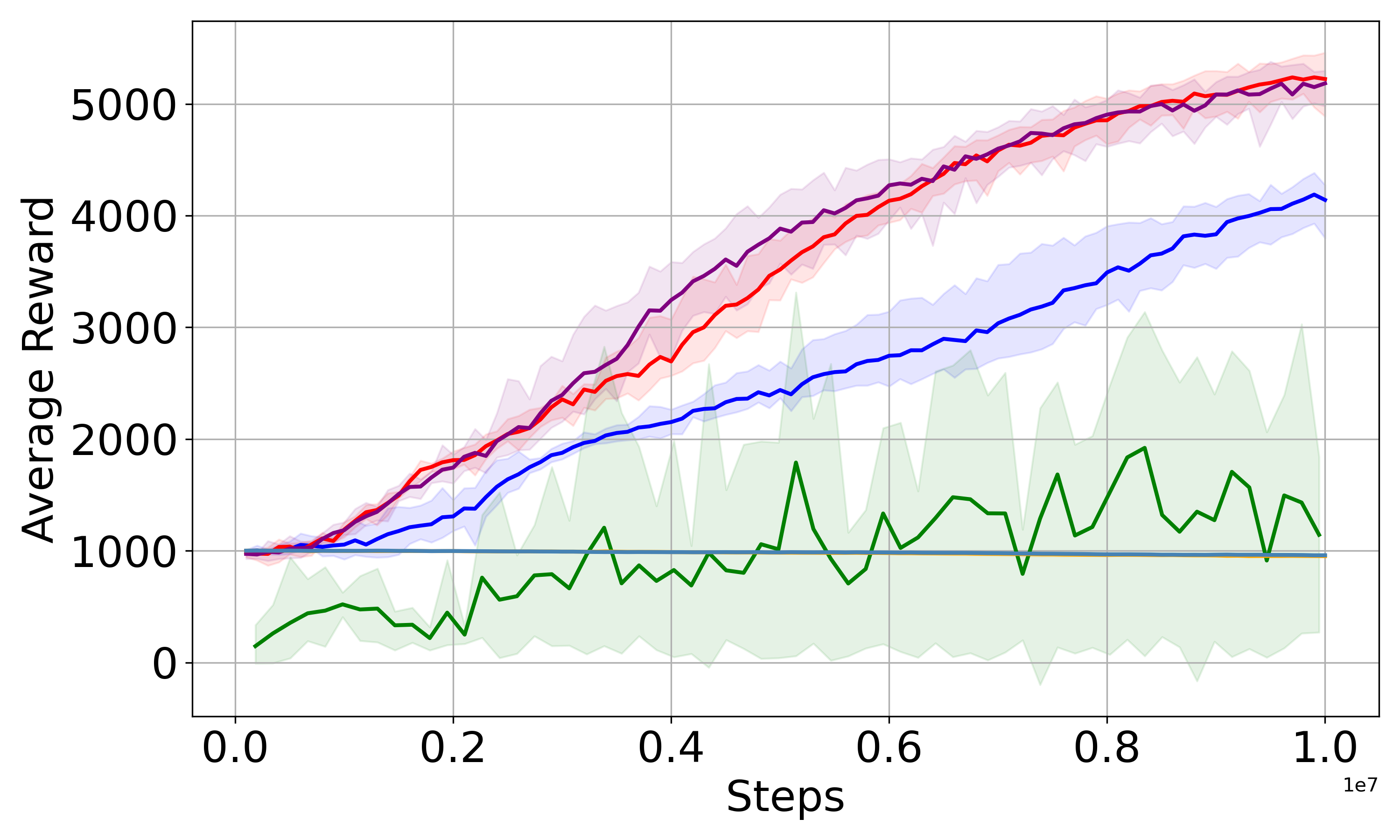}
            \label{fig:fig2}
        } &
        \subfloat[8x1-Agent Ant]{%
            \includegraphics[width=0.22\textwidth]{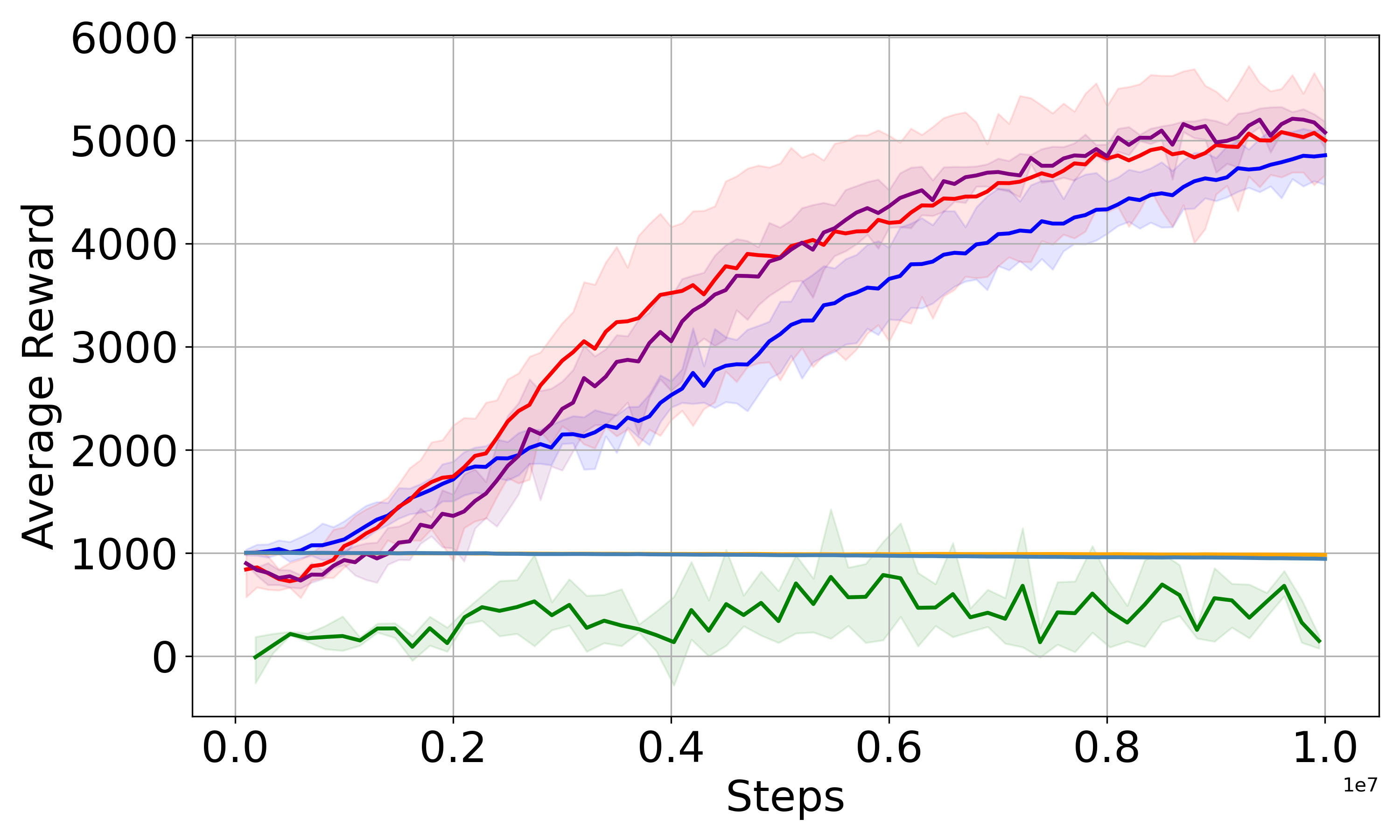}
            \label{fig:fig3}
        } &
        \multirow{3}{*}[2em]{%
            \subfloat[KL divergence over time for each agent across
methods]{
            \includegraphics[width=0.23\textwidth]{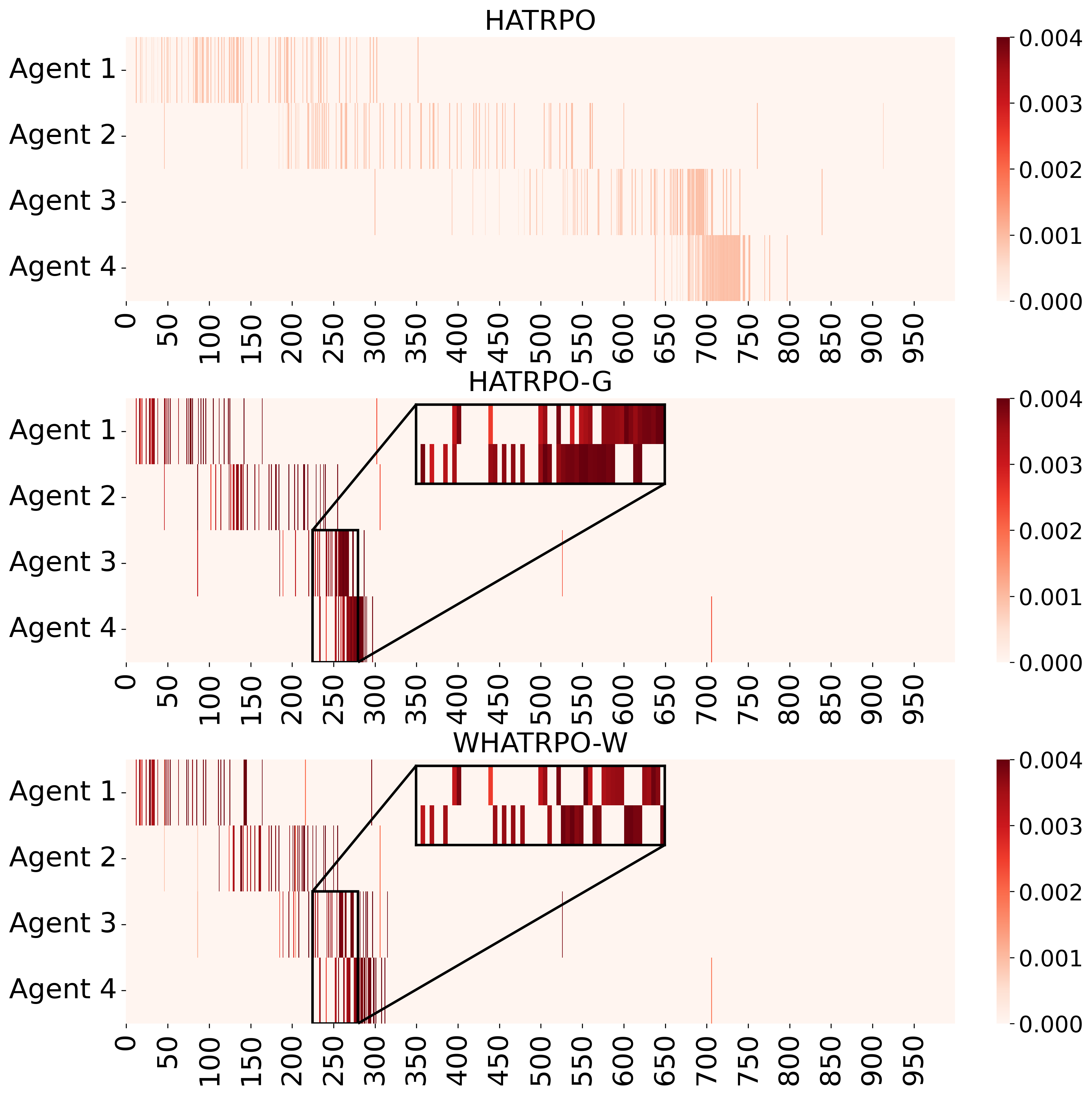}
            \label{fig:kl_heatmap}
            }
        } \\

        \subfloat[2x3-Agent HalfCheetah]{%
            \includegraphics[width=0.22\textwidth]{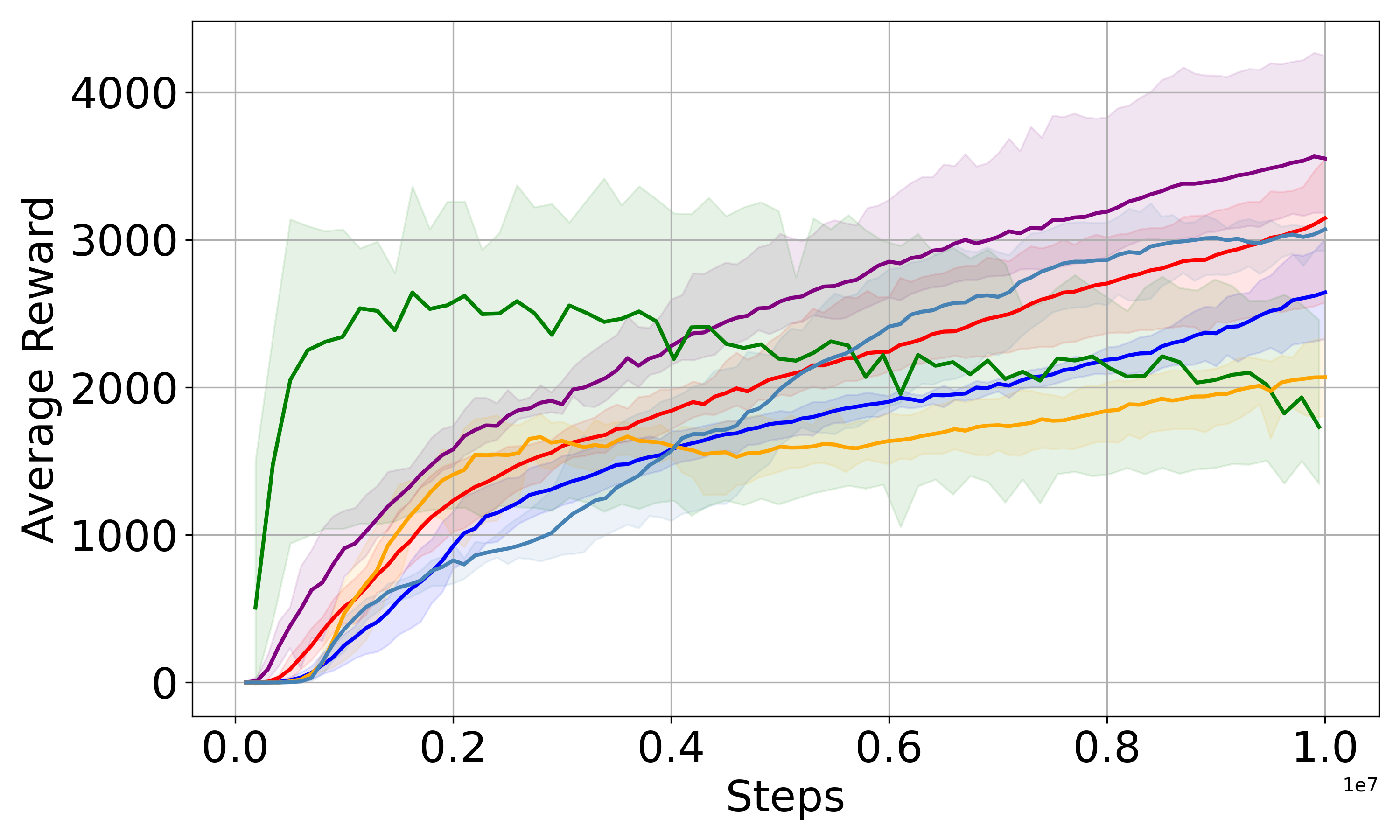}
            \label{fig:fig4}
        } &
        \subfloat[3x2-Agent HalfCheetah]{%
            \includegraphics[width=0.22\textwidth]{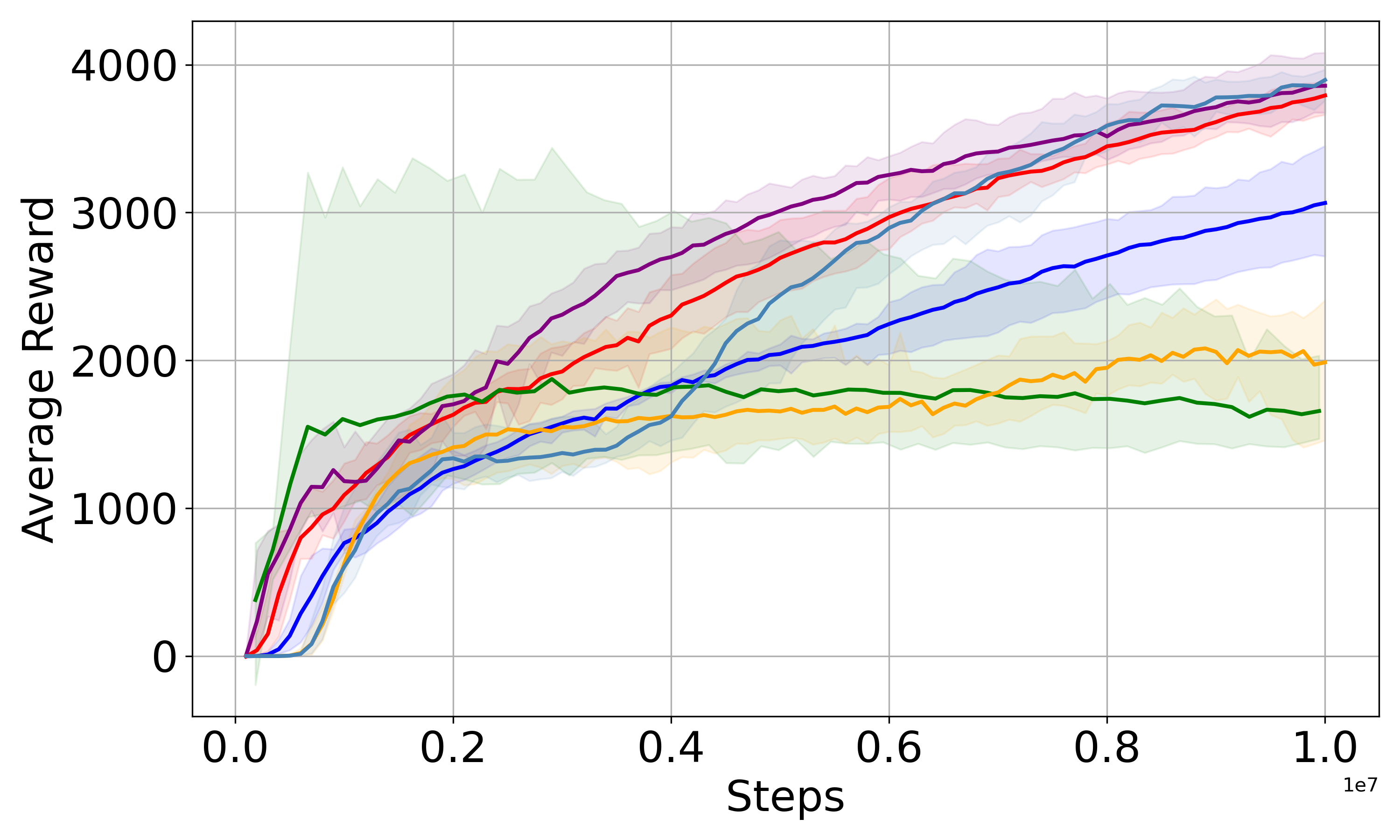}
            \label{fig:fig5}
        } &
        \subfloat[6x1-Agent HalfCheetah]{%
            \includegraphics[width=0.22\textwidth]{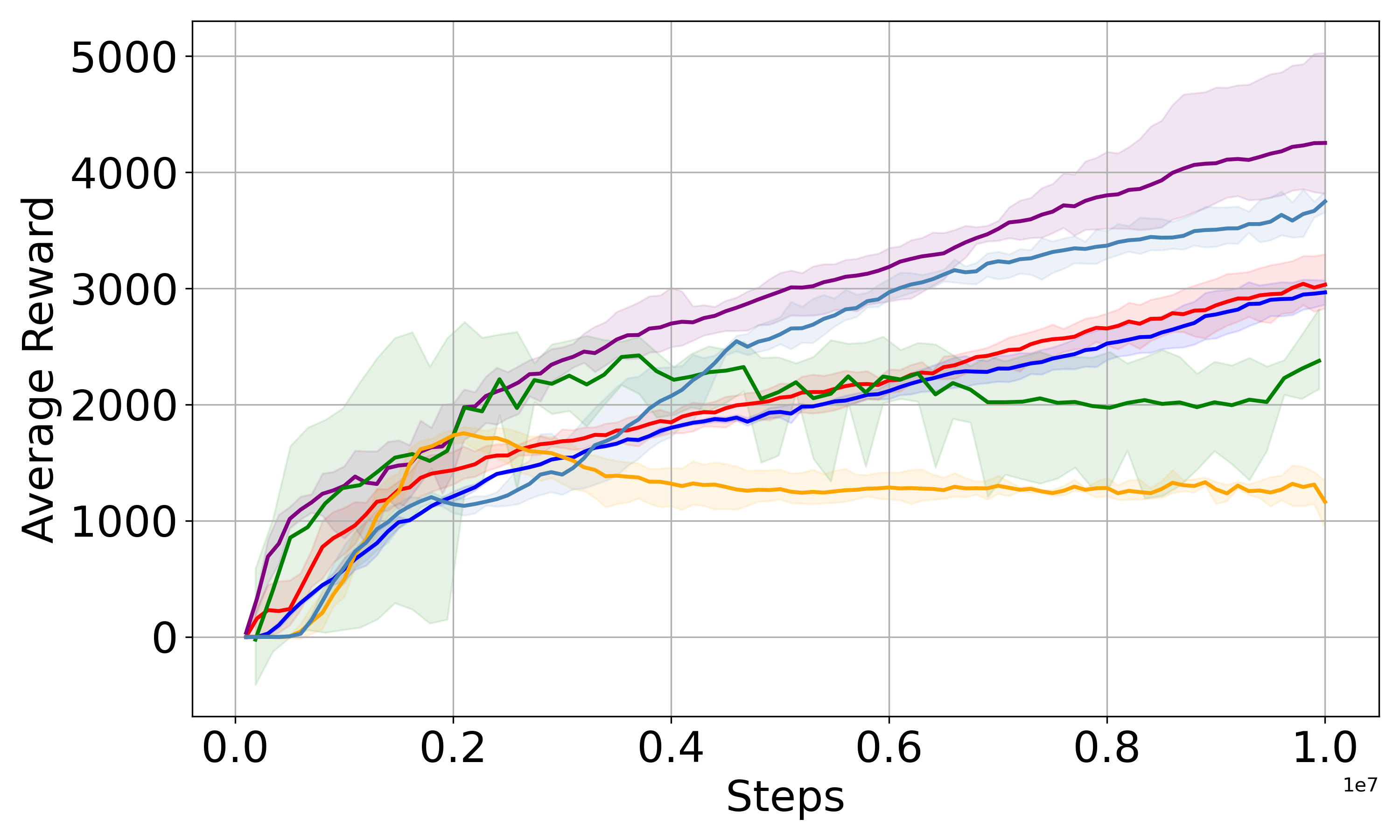}
            \label{fig:fig6}
        } & \\%

        \subfloat[3x1-Agent Hopper]{%
            \includegraphics[width=0.22\textwidth]{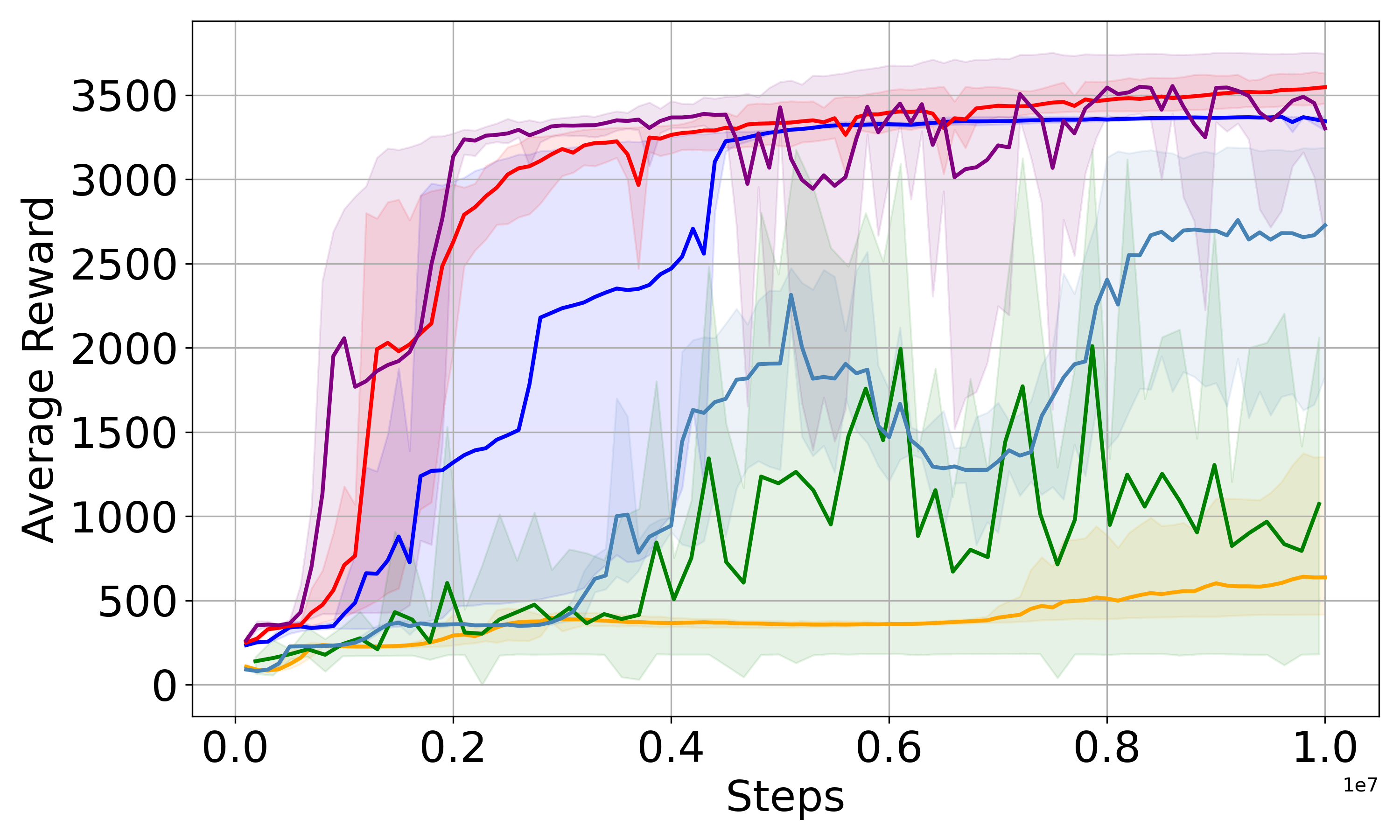}
            \label{fig:fig7}
        } &
        \subfloat[3x2-Agent Walker]{%
            \includegraphics[width=0.22\textwidth]{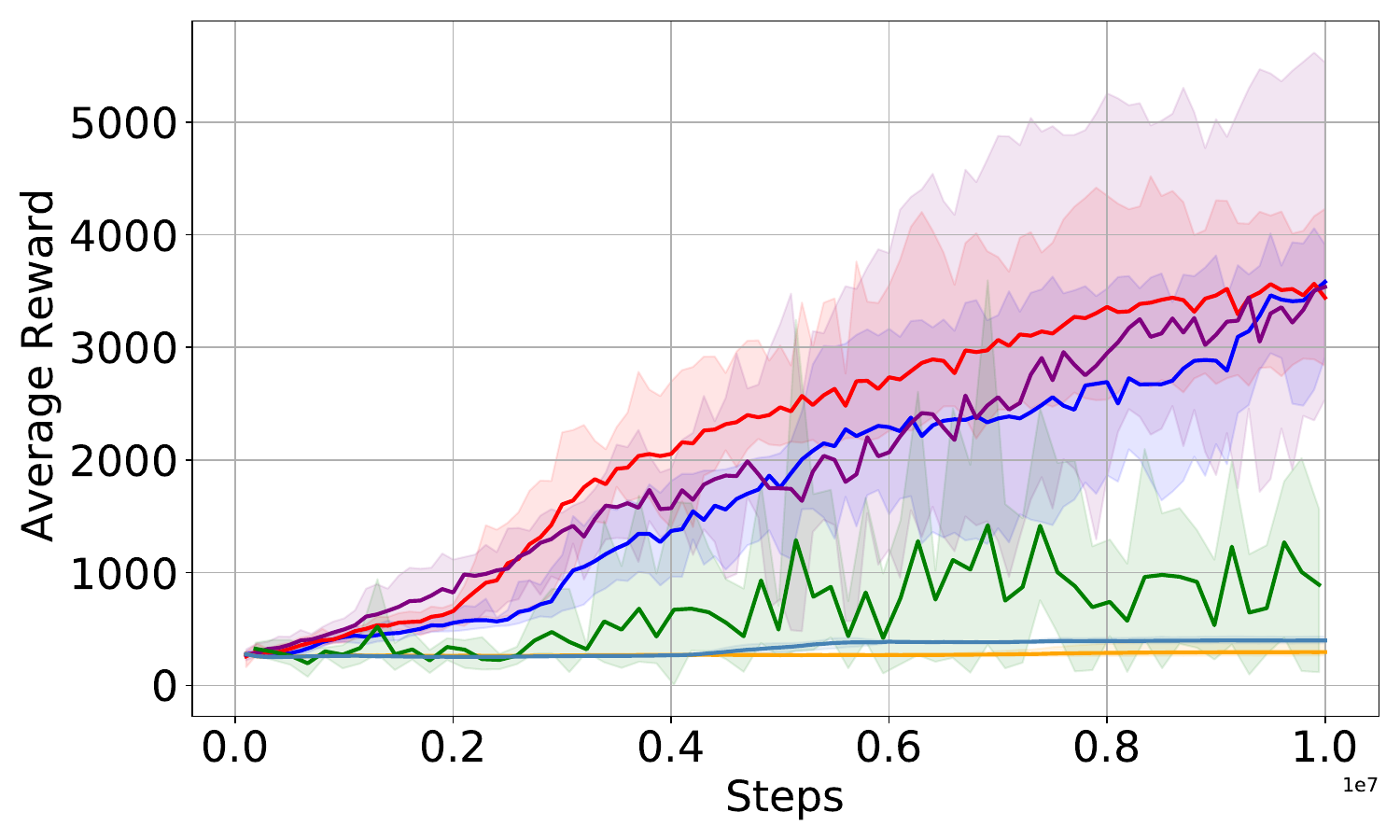}
            \label{fig:fig8}
        } &
        \subfloat[6x1-Agent Walker]{%
            \includegraphics[width=0.22\textwidth]{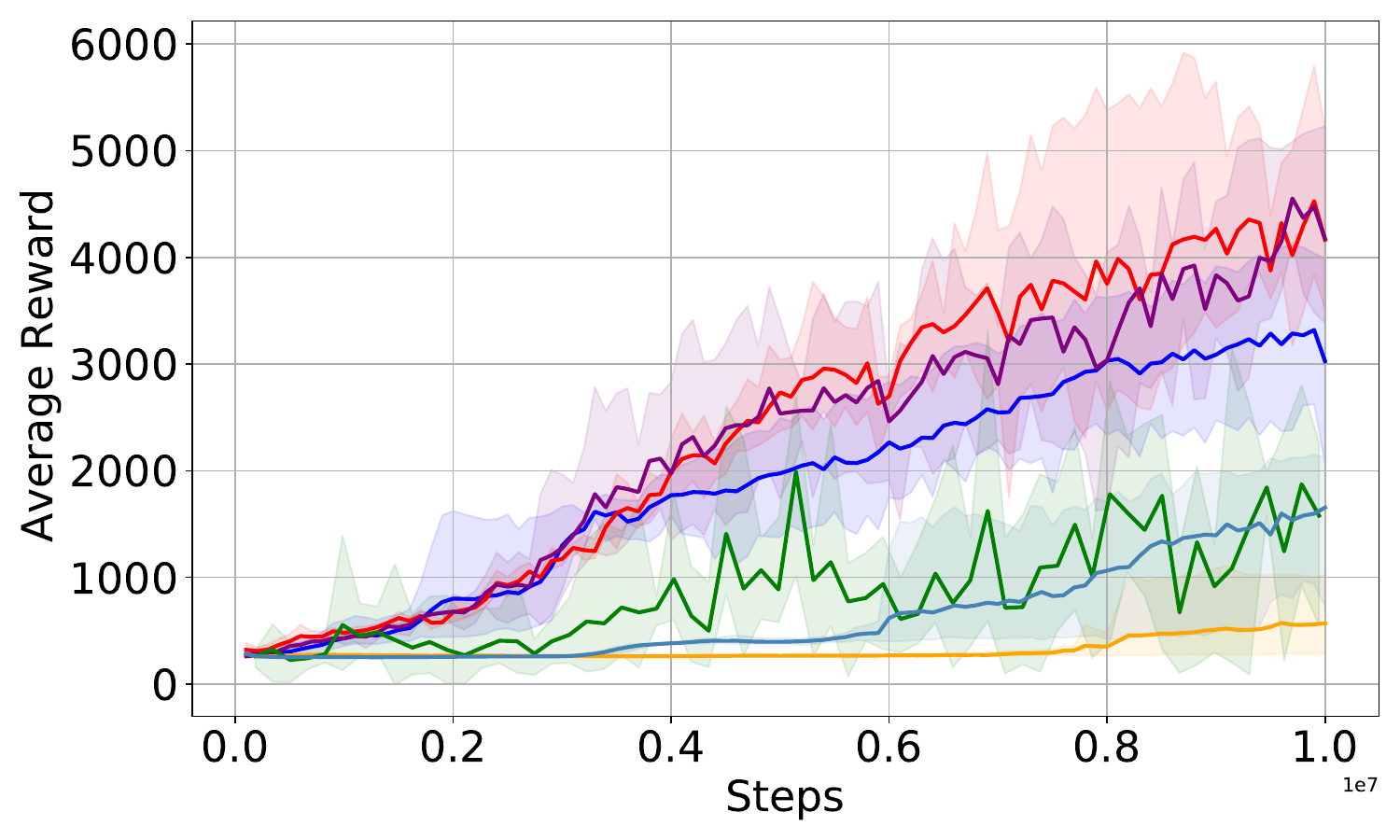}
            \label{fig:fig9}
        } &
        \\
    \end{tabular}

    \caption{(a)-(c), (e)-(g), and (h)-(j): Performance comparison on multiple Multi-Agent MuJoCo tasks.  (d): KL divergence over time for each agent across methods. Each heatmap shows the per-agent KL divergence at each policy update iteration.
        HATRPO (top) applies a uniform threshold to all agents, resulting in consistently low and evenly spread KL values over time.
        In contrast, HATRPO-G (middle) and HATRPO-W (bottom) adaptively allocate KL thresholds based on agent-specific learning signals.}
    \vspace{-10pt}
    \label{fig:performance_mojoco}
\end{figure*}

~\\
Our algorithm consistently outperforms or matches strong baselines across diverse environments under a fixed total KL divergence threshold shown in Figure \ref{fig:performance_mojoco} . Specifically, HATRPO-G improves final performance by 25.2\%, while HATRPO-W achieves a 22.5\% gain over the original HATRPO. However, HATRPO-G exhibits higher variability, with a standard deviation 39\% greater than that of HATRPO-W. This reflects the broader generalization capabilities of our methods HATRPO-W. In both the matrix game and the differential game, our approach converges significantly faster and successfully escapes suboptimal local minima with reward plateaus, where HATRPO remains trapped.  On Multi-Agent MuJoCo tasks, as shown in Fig. \ref{fig:performance_mojoco}, particularly Ant and HalfCheetah, our method achieves notably higher returns. These results underscore the effectiveness and robustness of our KL-adaptive strategy across both discrete and continuous domains.

\subsubsection{(b) Can adaptive KL allocation lead to more effective and structured policy updates, especially in heterogeneous or imbalanced agent settings?}
~\\
We provide further analysis of policy behavior in the matrix game with 10 agents and differential game (see Figure~\ref{fig:Matrix_Policy} and Figure~\ref{fig:Gaussian_Policy}). In the matrix game, the reward structure depends on the actions of earlier-indexed agents, leading to asymmetries in advantage estimation. As a result, agents earlier in the update sequence tend to exhibit greater policy changes, while later agents often show minimal adaptation. Allocating more KL threshold to agents with higher advantage values—typically those in earlier positions—facilitates faster group convergence.

In contrast, the differential Gaussian game illustrates a case where standard HATRPO becomes stuck at a local optimum. Here, equal KL allocation prevents exploration beyond the local peak at (2, 1), especially for Agent 1. Our adaptive variants (HATRPO-G/W), which dynamically reassign KL constraints, enable Agent 1 to pursue larger updates and ultimately escape to the global optimum near (5, 5), demonstrating the benefits of asymmetric KL allocation for structured exploration and escape from local minima.

\subsubsection{(c) Does our method accelerate convergence by prioritizing policy updates for high-advantage agents?}
~\\
We further evaluate the convergence efficiency of each method in a 4-agent matrix game by measuring the number of steps required to reach 99\% of the maximum achievable reward. As shown in Figure~\ref{fig:Matrix_steps}, HATRPO consistently requires more iterations to reach optimality, especially under tighter KL constraints (smaller $\delta$ values). This performance gap underscores the inefficiencies of uniform KL distribution in environments with asymmetric advantage distributions. In contrast, both HATRPO-G and HATRPO-W exhibit significantly faster convergence. 

\subsubsection{(d) How well does the allocated KL threshold reflect each agent’s contribution? }
\begin{figure}[ht]
    \centering
     \subfloat[]{%
        \includegraphics[width=0.25\textwidth]{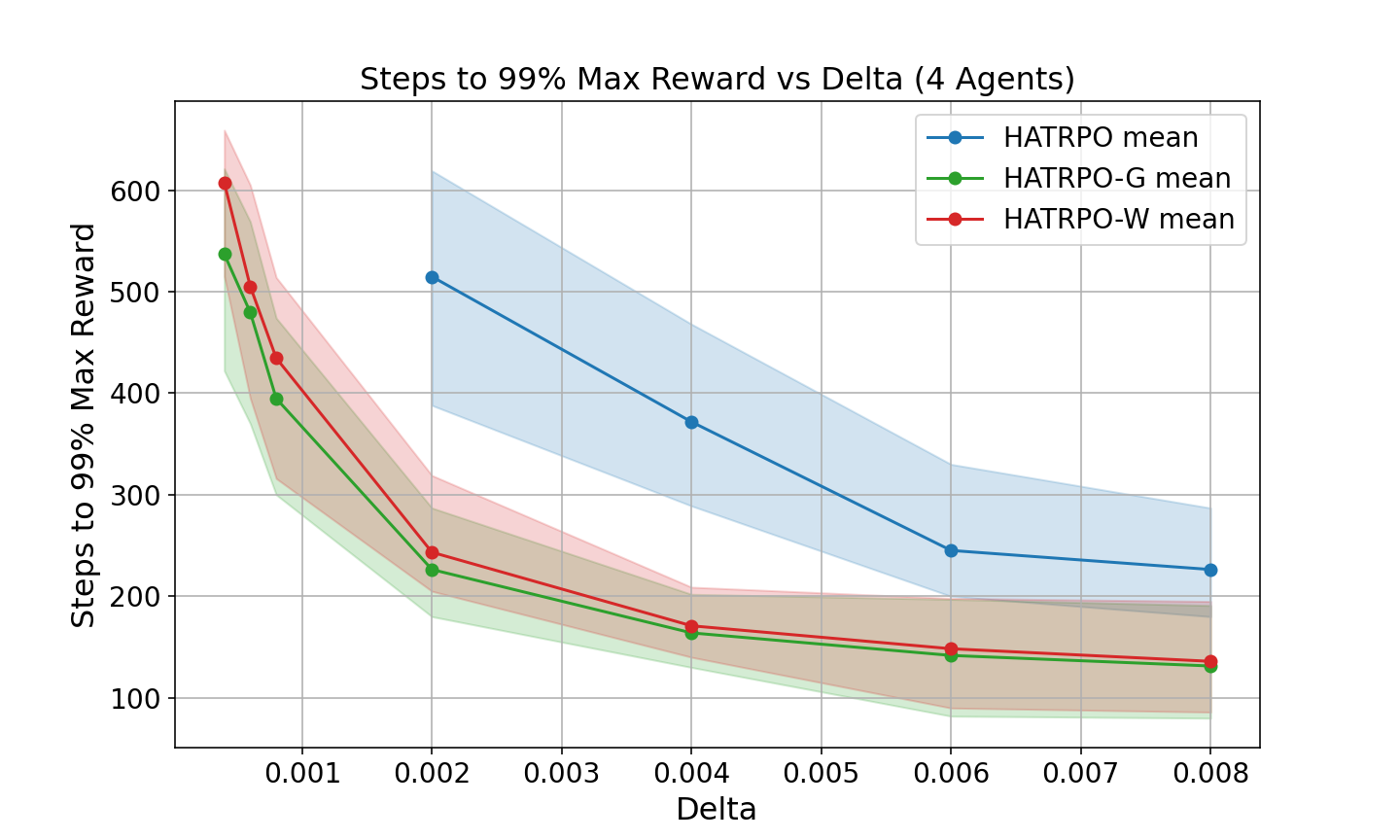}
        \label{fig:Matrix_steps}
    }
    \subfloat[]{%
        \includegraphics[width=0.21\textwidth]{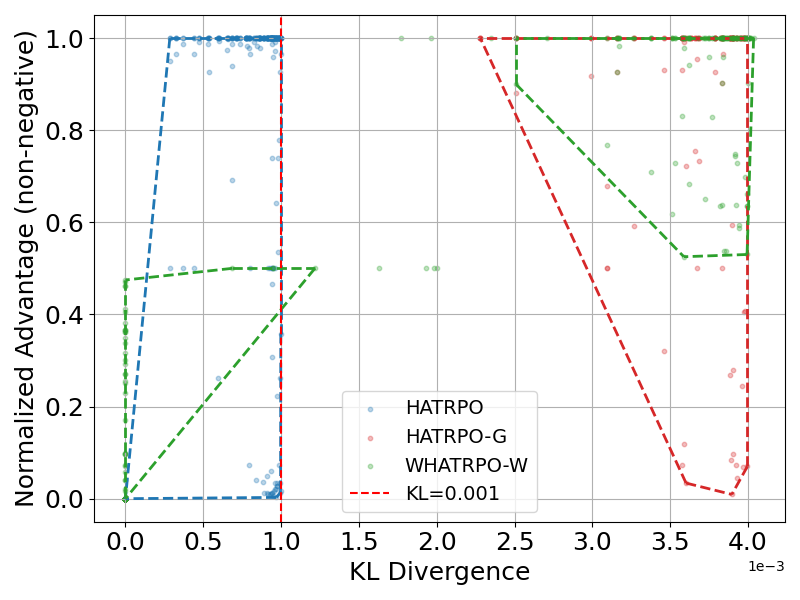}
        \label{fig:Matrix_Adv}
    }
    \caption{(a) Number of steps required to reach 99\% of the maximum reward as a function of $\delta$ in the 4-agent matrix game. (b) {Normalized advantage vs.\ KL divergence for all methods with convex boundaries.}
        Each point represents an agent's KL-advantage pair.}
        \vspace{-10 pt}
    \label{fig:Matrix_steps_Adv}
\end{figure}

~\\
The allocated KL threshold closely reflects each agent’s contribution as indicated by its advantage function. As shown in Figure~\ref{fig:Matrix_Adv}, both HATRPO-G and HATRPO-W exhibit a strong correlation between KL divergence and normalized advantage. HATRPO-W allocates more KL to agents with the highest individual advantage. In contrast, HATRPO-G distributes KL more collectively, favoring groups of agents with consistently high advantage. This alignment between KL allocation and advantage demonstrates that our adaptive methods effectively prioritize impactful updates, in contrast to HATRPO’s uniform strategy which neglects agent-specific learning potential.

\subsubsection{(e) Can our method overcome the bottlenecks of uniform KL constraints and improve joint policy optimization?}
~\\
Yes, our method overcomes the bottlenecks of uniform KL constraints and leads to improved joint policy optimization. As shown in Figure~\ref{fig:kl_heatmap}, HATRPO allocates low and uniformly distributed KL values across agents and time, limiting its ability to prioritize critical updates. In contrast, both HATRPO-W and HATRPO-G exhibit structured and adaptive KL allocation. Notably, agent updates follow a sequential order (from agent 1 to agent 4), and the adaptive methods leverage this structure to focus learning more effectively. HATRPO-W updates agents in an alternating pattern—e.g., agent 3 → agent 4 → agent 3—allowing for iterative refinement, while HATRPO-G concentrates KL budget on a subset of high-advantage agents, enabling parallel but selective updates. This dynamic allocation enables more expressive and efficient joint policy optimization, particularly in heterogeneous-agent settings.

\section{Conclusion}

We introduced HATRPO-G and HATRPO-W, two adaptive extensions of HATRPO that address the inefficiencies of uniform KL divergence constraints in multi-agent reinforcement learning. By reallocating the total KL threshold based on agents’ advantage signals, our methods enable more targeted policy updates that improve both learning speed and final performance. Experiments across matrix games, differential games, and multi-agent MuJoCo environments demonstrate that our adaptive algorithms converge faster and better utilize the available policy update threshold, particularly in settings with heterogeneous roles or imbalanced agent importance. These results suggest that our methods improve overall performance, promote more structured learning updates, and accelerate convergence, all while operating under a fixed total KL divergence budget. The learned KL threshold allocation effectively captures variations in agent advantage and overcomes the limitations of fixed-threshold designs, leading to more coordinated and efficient multi-agent learning.



\section{Appendix}

\subsection{Computing Infrastructure}

All experiments were conducted on a machine with Red Hat Enterprise Linux 8.10 (Ootpa), equipped with 4 CPUs, 16/32 GB of RAM, and eight NVIDIA GeForce GTX 2080 Ti GPUs. GPUs were accelerated using CUDA and cuDNN. This hardware and software setup was used consistently across all benchmarks. To ensure reproducibility, we used 5 random seeds for experiments in the Multi-Agent MuJoCo environment and 10 seeds for both the differential game and matrix game settings. The method for setting seeds is explicitly implemented in our code using functions such as \texttt{np.random.seed(seed)}, allowing others to replicate the experiments with the same random initialization.

\subsection{Detailed Description of MARL Benchmark}
     \noindent \textbf{Matrix Game: Coordination Under Sparse Reward.} We consider a synthetic matrix game with $N$ agents, each having a binary action space $\{0, 1\}$ and a single shared state. The reward function is asymmetric and sparse, yielding the maximum reward (1.5) only when all agents simultaneously choose action 1. This setup enables us to study how KL allocation affects convergence speed in the presence of non-uniform advantage distributions.

    \noindent \textbf{Differential Game: Escaping Local Optima.} We design a two-player differential game where agents act in a continuous space $[0, 7]$ using Gaussian policies. The reward surface includes both a local and a global optimum, requiring agents to explore asymmetrically to achieve optimality. This environment tests the ability of adaptive KL strategies to escape local optima.

\begin{figure}[ht]
    \centering
    \includegraphics[width=0.9\linewidth]{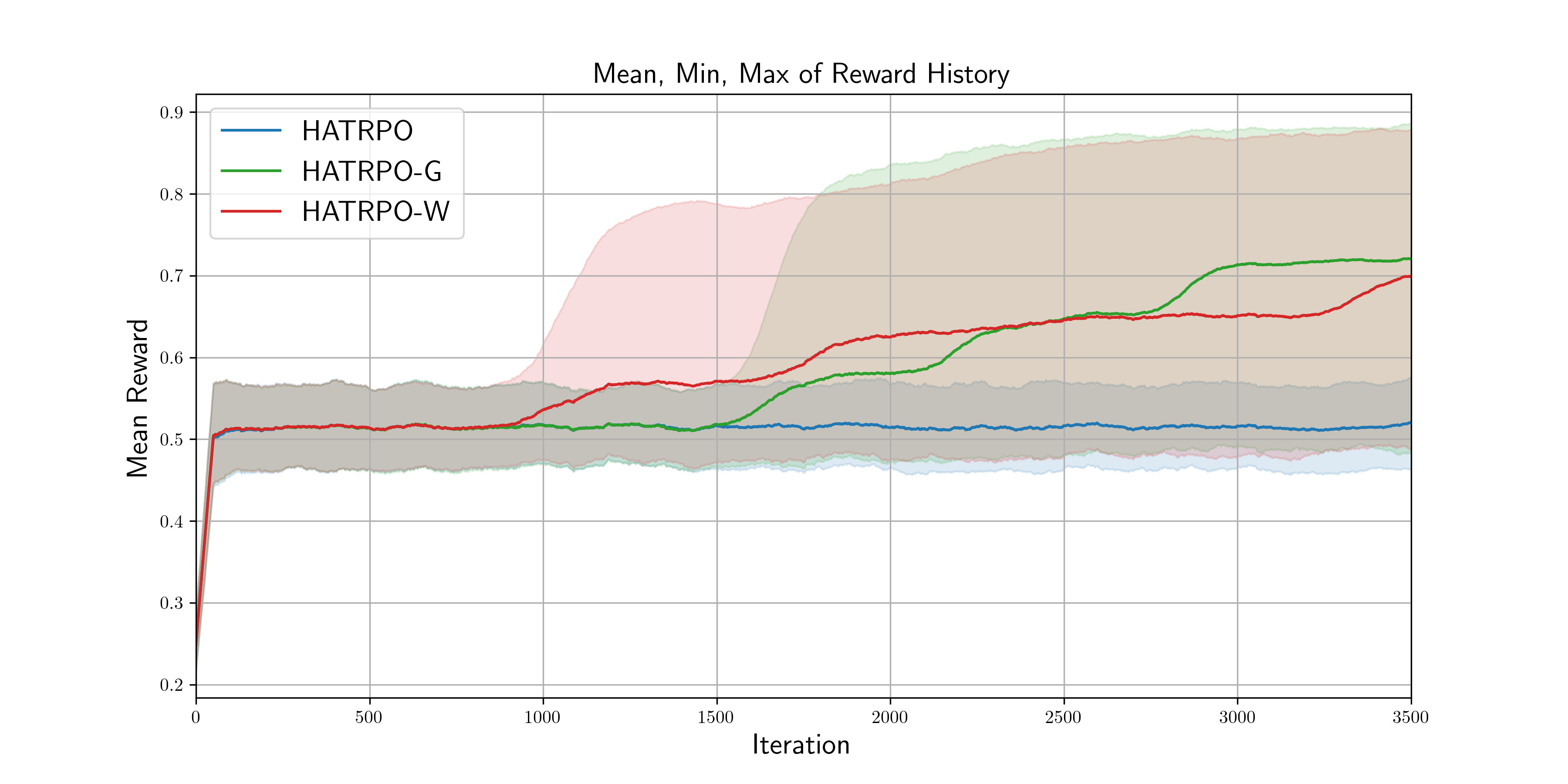}
    \caption{Comparison of average reward over training iterations for HATRPO (blue), HATRPO-G (green), and HATRPO-W (red) in a continuous control setting. Both HATRPO-G and HATRPO-W demonstrate improved learning efficiency and higher final performance compared to the baseline HATRPO, with HATRPO-W showing faster convergence in early stages.
    }
    \label{fig:Guassian_reward}
\end{figure}

     \noindent \textbf{Multi-Agent MuJoCo: Realistic Heterogeneous Agents.} In this benchmark, agents correspond to physical components of a robot (e.g., Ant), introducing inherent heterogeneity in roles and dynamics. We evaluate performance across 12 MuJoCo tasks, assessing the scalability and adaptability of our methods in realistic settings. 

\subsection{Performance in Two-play Differential Game}
Figure~\ref{fig:Guassian_reward} illustrates the reward learning curves for a two-player differential game with a Gaussian reward function. The baseline HATRPO method exhibits early convergence to a suboptimal local optimum, failing to explore regions of higher reward. In contrast, both HATRPO-G and HATRPO-W demonstrate the ability to escape this local minimum and achieve substantially higher rewards. These results underscore the importance of adaptive KL allocation in enabling more effective policy updates and robust performance.

\subsection{Hyperparameters}
\begin{table}[h]
\centering
\begin{tabular}{|l|l|}
\hline
\textbf{Hyperparameter} & \textbf{Value} \\
\hline
Number of iterations             & 4000 \\
Batch size                       & 20 \\
Number of evaluation episodes    & 1000 \\
Initial mean $(\mu_1, \mu_2)$     & (1.0, 1.0) \\
Standard deviation ($\sigma$)   & 1.15 \\
Critic lr                   & 0.2 \\
tolerance of HATRPO-W & 0.01 \\
Constant of HATRPO-G & 1e-4 \\
\hline
\end{tabular}
\caption{Hyperparameters used for HATRPO, HATRPO-G, and WHATRPO in differential game simulations.}
\label{tab:hatrpo_hyperparams2}
\end{table}

\begin{table}[h]
\centering
\begin{tabular}{|l l|}
\hline
\textbf{hyperparameters} & \textbf{value} \\
\hline
critic lr       & 5e-3   \\
optimizer         & Adam  \\
num mini-batch  & 1     \\
gamma           & 0.99   \\
optim eps         & 1e-5   \\
batch size       & 4000  \\
gain            & 0.01   \\
hidden layer      & 3    \\
training threads & 4     \\
std y coef      & 0.5    \\
actor network     & mlp   \\
rollout threads  & 20     \\
std x coef      & 1      \\
max grad norm     & 10    \\
episode length   & 200  \\
activation      & ReLU   \\
hidden layer dim  & 128   \\
eval episode     & 40    \\
actor lr    & 5e-6  \\
ppo epoch    & 5 \\
ppo clip & 0.2 \\
\hline
\end{tabular}
\caption{Hyperparameters for MAPPO and HAPPO used in the Multi-Agent MuJoCo domain.}
\label{tab:MAPPO1}
\end{table}

\begin{table}[h]
\centering
\begin{tabular}{|l l|}
\hline
\textbf{hyperparameters} & \textbf{value} \\
\hline
critic lr       & 5e-4   \\
optimizer         & Adam  \\
num mini-batch  & 1     \\
gamma           & 0.99   \\
optim eps         & 1e-5   \\
batch size       & 4000  \\
gain            & 0.01   \\
hidden layer      & 3    \\
training threads & 4     \\
std y coef      & 0.5    \\
actor network     & mlp   \\
rollout threads  & 20     \\
std x coef      & 1      \\
max grad norm     & 10    \\
episode length   & 200  \\
activation      & ReLU   \\
hidden layer dim  & 128   \\
eval episode     & 40    \\
accept ratio     & 0.5   \\
tolerance of HATRPO-W & 0.01 \\
Constant of HATRPO-G & 1e-4 \\

\hline
\end{tabular}
\caption{Hyperparameters for HATRPO, HATRPO-W, and HATRPO-G used across in the Multi-Agent MuJoCo domain.}
\label{tab:MAPPO2}
\end{table}

\begin{table}[h]
\centering
\begin{tabular}{|l l|}
\hline
\textbf{Task} & \textbf{KL-threshold} \\
\hline
Ant (2x4)              & 5e-4 \\
Ant (4x2)              & 5e-4 \\
Ant (8x1)              & 5e-4 \\
HalfCheetah (2x3)      & 5e-5 \\
HalfCheetah (3x2)      & 1e-4 \\
HalfCheetah (6x1)      & 5e-5 \\
Hopper (3x1)           & 5e-5 \\
Walker (2x3)           & 1e-3 \\
Walker (3x2)           & 5e-4 \\
\hline
\end{tabular}
\caption{KL-threshold values used for HATRPO, HATRPO-W, HATRPO-G across environments in the Multi-Agent MuJoCo domain.}
\label{tab:kl_threshold_singlecol}
\end{table}

\begin{table}[h]
\centering
\begin{tabular}{|l l|}
\hline
\textbf{Hyperparameter} & \textbf{Value} \\
\hline
actor lr               & 1e-3      \\
critic lr              & 1e-3      \\
gamma                  & 0.99      \\
tau                    & 5e-2      \\
start-timesteps        & 25000     \\
optimizer              & Adam      \\
exploration noise      & 0.1       \\
step-per-epoch         & 50000     \\
step-per-collector     & 2000      \\
update-per-step        & 0.05      \\
hidden sizes           & 256-256   \\
buffer size            & 1e6       \\
batch size             & 200       \\
training num           & 16        \\
test num               & 10        \\
n-step                 & 1         \\
epoch                  & 200       \\
episode length         & 1000       \\
\hline
\end{tabular}
\caption{Selected hyperparameters used for MADDPG in the Multi-Agent MuJoCo domain.}
\label{tab:maddpg_selected_hyperparams}
\end{table}

\begin{table}[h]
\centering
\begin{tabular}{|l|l|}
\hline
\textbf{Hyperparameter} & \textbf{Value} \\
\hline
Number of iterations             & 1000 \\
Batch size                       & 20 \\
Number of evaluation episodes    & 100 \\
Initial action probability       & (0.99, 0.01) \\
tolerance of HATRPO-W & 0.01 \\
Constant of HATRPO-G & 1e-4 \\
\hline
\end{tabular}
\caption{Hyperparameters used for HATRPO, HATRPO-G, and WHATRPO in matrix game simulations.}
\label{tab:hatrpo_hyperparams1}
\end{table}

\bibliography{main}

\begin{thebibliography}{33}
\providecommand{\natexlab}[1]{#1}

\bibitem[{Bhalla, Ganapathi~Subramanian, and Crowley(2020)}]{bhalla2020deep}
Bhalla, S.; Ganapathi~Subramanian, S.; and Crowley, M. 2020.
\newblock Deep multi agent reinforcement learning for autonomous driving.
\newblock In \emph{Canadian Conference on Artificial Intelligence}, 67--78. Springer.

\bibitem[{Canese et~al.(2021)Canese, Cardarilli, Di~Nunzio, Fazzolari, Giardino, Re, and Span{\`o}}]{canese2021multi}
Canese, L.; Cardarilli, G.~C.; Di~Nunzio, L.; Fazzolari, R.; Giardino, D.; Re, M.; and Span{\`o}, S. 2021.
\newblock Multi-agent reinforcement learning: A review of challenges and applications.
\newblock \emph{Applied Sciences}, 11(11): 4948.

\bibitem[{De~Witt et~al.(2020)De~Witt, Gupta, Makoviichuk, Makoviychuk, Torr, Sun, and Whiteson}]{de2020independent}
De~Witt, C.~S.; Gupta, T.; Makoviichuk, D.; Makoviychuk, V.; Torr, P.~H.; Sun, M.; and Whiteson, S. 2020.
\newblock Is independent learning all you need in the starcraft multi-agent challenge?
\newblock \emph{arXiv preprint arXiv:2011.09533}.

\bibitem[{Dou et~al.(2024)Dou, Dang, Luan, and Chen}]{dou2024measuring}
Dou, H.; Dang, L.; Luan, Z.; and Chen, B. 2024.
\newblock Measuring mutual policy divergence for multi-agent sequential exploration.
\newblock \emph{Advances in Neural Information Processing Systems}, 37: 76265--76288.

\bibitem[{Feng et~al.(2023)Feng, Xing, Zhang, and Pan}]{feng2023fp3o}
Feng, L.; Xing, D.; Zhang, J.; and Pan, G. 2023.
\newblock FP3O: Enabling proximal policy optimization in multi-agent cooperation with parameter-sharing versatility.
\newblock \emph{arXiv preprint arXiv:2310.05053}.

\bibitem[{Gu et~al.(2023)Gu, Kuba, Chen, Du, Yang, Knoll, and Yang}]{gu2023safe}
Gu, S.; Kuba, J.~G.; Chen, Y.; Du, Y.; Yang, L.; Knoll, A.; and Yang, Y. 2023.
\newblock Safe multi-agent reinforcement learning for multi-robot control.
\newblock \emph{Artificial Intelligence}, 319: 103905.

\bibitem[{{He} et~al.(2013){He}, {Zhao}, {Zhou}, and {Niu}}]{2013ITWC...12.3637H}
{He}, P.; {Zhao}, L.; {Zhou}, S.; and {Niu}, Z. 2013.
\newblock {Water-Filling: A Geometric Approach and its Application to Solve Generalized Radio Resource Allocation Problems}.
\newblock \emph{IEEE Transactions on Wireless Communications}, 12(7): 3637--3647.

\bibitem[{Kuba et~al.(2021)Kuba, Chen, Wen, Wen, Sun, Wang, and Yang}]{kuba2021trust}
Kuba, J.~G.; Chen, R.; Wen, M.; Wen, Y.; Sun, F.; Wang, J.; and Yang, Y. 2021.
\newblock Trust region policy optimisation in multi-agent reinforcement learning.
\newblock \emph{arXiv preprint arXiv:2109.11251}.

\bibitem[{Kuba et~al.(2022)Kuba, Feng, Ding, Dong, Wang, and Yang}]{kuba2022heterogeneous}
Kuba, J.~G.; Feng, X.; Ding, S.; Dong, H.; Wang, J.; and Yang, Y. 2022.
\newblock Heterogeneous-agent mirror learning: A continuum of solutions to cooperative marl.
\newblock \emph{arXiv preprint arXiv:2208.01682}.

\bibitem[{Liu and Liu(2024)}]{liu2024jointppo}
Liu, C.; and Liu, G. 2024.
\newblock JointPPO: diving deeper into the effectiveness of PPO in multi-agent reinforcement learning.
\newblock \emph{arXiv preprint arXiv:2404.11831}.

\bibitem[{Liu et~al.(2023)Liu, Zhong, Hu, Fu, Fu, Chang, and Yang}]{liu2023maximum}
Liu, J.; Zhong, Y.; Hu, S.; Fu, H.; Fu, Q.; Chang, X.; and Yang, Y. 2023.
\newblock Maximum entropy heterogeneous-agent reinforcement learning.
\newblock \emph{arXiv preprint arXiv:2306.10715}.

\bibitem[{Lowe et~al.(2017)Lowe, Wu, Tamar, Harb, Pieter~Abbeel, and Mordatch}]{lowe2017multi}
Lowe, R.; Wu, Y.~I.; Tamar, A.; Harb, J.; Pieter~Abbeel, O.; and Mordatch, I. 2017.
\newblock Multi-agent actor-critic for mixed cooperative-competitive environments.
\newblock \emph{Advances in neural information processing systems}, 30.

\bibitem[{Mnih et~al.(2015)Mnih, Kavukcuoglu, Silver, Rusu, Veness, Bellemare, Graves, Riedmiller, Fidjeland, Ostrovski et~al.}]{mnih2015human}
Mnih, V.; Kavukcuoglu, K.; Silver, D.; Rusu, A.~A.; Veness, J.; Bellemare, M.~G.; Graves, A.; Riedmiller, M.; Fidjeland, A.~K.; Ostrovski, G.; et~al. 2015.
\newblock Human-level control through deep reinforcement learning.
\newblock \emph{nature}, 518(7540): 529--533.

\bibitem[{Park and Moon(2022)}]{park2022multi}
Park, K.; and Moon, I. 2022.
\newblock Multi-agent deep reinforcement learning approach for EV charging scheduling in a smart grid.
\newblock \emph{Applied energy}, 328: 120111.

\bibitem[{Rashid et~al.(2020{\natexlab{a}})Rashid, Farquhar, Peng, and Whiteson}]{rashid2020weighted}
Rashid, T.; Farquhar, G.; Peng, B.; and Whiteson, S. 2020{\natexlab{a}}.
\newblock Weighted qmix: Expanding monotonic value function factorisation for deep multi-agent reinforcement learning.
\newblock \emph{Advances in neural information processing systems}, 33: 10199--10210.

\bibitem[{Rashid et~al.(2020{\natexlab{b}})Rashid, Samvelyan, De~Witt, Farquhar, Foerster, and Whiteson}]{rashid2020monotonic}
Rashid, T.; Samvelyan, M.; De~Witt, C.~S.; Farquhar, G.; Foerster, J.; and Whiteson, S. 2020{\natexlab{b}}.
\newblock Monotonic value function factorisation for deep multi-agent reinforcement learning.
\newblock \emph{Journal of Machine Learning Research}, 21(178): 1--51.

\bibitem[{Roesch et~al.(2020)Roesch, Linder, Zimmermann, Rudolf, Hohmann, and Reinhart}]{roesch2020smart}
Roesch, M.; Linder, C.; Zimmermann, R.; Rudolf, A.; Hohmann, A.; and Reinhart, G. 2020.
\newblock Smart grid for industry using multi-agent reinforcement learning.
\newblock \emph{Applied Sciences}, 10(19): 6900.

\bibitem[{Schulman et~al.(2015)Schulman, Levine, Abbeel, Jordan, and Moritz}]{schulman2015trust}
Schulman, J.; Levine, S.; Abbeel, P.; Jordan, M.; and Moritz, P. 2015.
\newblock Trust region policy optimization.
\newblock In \emph{International conference on machine learning}, 1889--1897. PMLR.

\bibitem[{Schulman et~al.(2017)Schulman, Wolski, Dhariwal, Radford, and Klimov}]{schulman2017proximal}
Schulman, J.; Wolski, F.; Dhariwal, P.; Radford, A.; and Klimov, O. 2017.
\newblock Proximal policy optimization algorithms.
\newblock \emph{arXiv preprint arXiv:1707.06347}.

\bibitem[{Shao et~al.(2024)Shao, Wang, Zhu, Xu, Song, Bi, Zhang, Zhang, Li, Wu et~al.}]{shao2024deepseekmath}
Shao, Z.; Wang, P.; Zhu, Q.; Xu, R.; Song, J.; Bi, X.; Zhang, H.; Zhang, M.; Li, Y.; Wu, Y.; et~al. 2024.
\newblock Deepseekmath: Pushing the limits of mathematical reasoning in open language models.
\newblock \emph{arXiv preprint arXiv:2402.03300}.

\bibitem[{Son et~al.(2020)Son, Ahn, Reyes, Shin, and Yi}]{son2020qtran++}
Son, K.; Ahn, S.; Reyes, R.~D.; Shin, J.; and Yi, Y. 2020.
\newblock QTRAN++: Improved value transformation for cooperative multi-agent reinforcement learning.
\newblock \emph{arXiv preprint arXiv:2006.12010}.

\bibitem[{Son et~al.(2019)Son, Kim, Kang, Hostallero, and Yi}]{son2019qtran}
Son, K.; Kim, D.; Kang, W.~J.; Hostallero, D.~E.; and Yi, Y. 2019.
\newblock Qtran: Learning to factorize with transformation for cooperative multi-agent reinforcement learning.
\newblock In \emph{International conference on machine learning}, 5887--5896. PMLR.

\bibitem[{Tao et~al.(2025)Tao, Xinhao, Cheng, Yulin, Qinghan, and Hongzhe}]{tao2025tsppo}
Tao, Y.; Xinhao, S.; Cheng, X.; Yulin, Y.; Qinghan, Z.; and Hongzhe, L. 2025.
\newblock TSPPO: Transformer-Based Sequential Proximal Policy Optimization for Multi-Agent Systems.
\newblock \emph{Preprint at Research Square}.

\bibitem[{Wang et~al.(2023)Wang, Tian, Wan, Wen, Wang, and Zhang}]{wang2023order}
Wang, X.; Tian, Z.; Wan, Z.; Wen, Y.; Wang, J.; and Zhang, W. 2023.
\newblock Order matters: Agent-by-agent policy optimization.
\newblock \emph{arXiv preprint arXiv:2302.06205}.

\bibitem[{Wen et~al.(2022)Wen, Chen, Yang, Li, Tian, Chen, and Wang}]{wen2022game}
Wen, Y.; Chen, H.; Yang, Y.; Li, M.; Tian, Z.; Chen, X.; and Wang, J. 2022.
\newblock A game-theoretic approach to multi-agent trust region optimization.
\newblock In \emph{International conference on distributed artificial intelligence}, 74--87. Springer.

\bibitem[{Wu et~al.(2023)Wu, Chandra, Guan, Bedi, and Manocha}]{wu2023intent}
Wu, X.; Chandra, R.; Guan, T.; Bedi, A.; and Manocha, D. 2023.
\newblock Intent-aware planning in heterogeneous traffic via distributed multi-agent reinforcement learning.
\newblock In \emph{Conference on Robot Learning}, 446--477. PMLR.

\bibitem[{Yu et~al.(2022)Yu, Velu, Vinitsky, Gao, Wang, Bayen, and Wu}]{yu2022surprising}
Yu, C.; Velu, A.; Vinitsky, E.; Gao, J.; Wang, Y.; Bayen, A.; and Wu, Y. 2022.
\newblock The surprising effectiveness of ppo in cooperative multi-agent games.
\newblock \emph{Advances in neural information processing systems}, 35: 24611--24624.

\bibitem[{Yu and Cioffi(2006)}]{4151217}
Yu, D.~D.; and Cioffi, J.~M. 2006.
\newblock SPC10-2: Iterative Water-filling for Optimal Resource Allocation in OFDM Multiple-Access and Broadcast Channels.
\newblock In \emph{IEEE Globecom 2006}, 1--5.

\bibitem[{Yu, Huang, and Chang(2021)}]{yu2021optimizing}
Yu, T.; Huang, J.; and Chang, Q. 2021.
\newblock Optimizing task scheduling in human-robot collaboration with deep multi-agent reinforcement learning.
\newblock \emph{Journal of Manufacturing Systems}, 60: 487--499.

\bibitem[{Zhang et~al.(2024)Zhang, Mao, Li, Xu, Li, Zhao, and Fan}]{zhang2024sequential}
Zhang, B.; Mao, H.; Li, L.; Xu, Z.; Li, D.; Zhao, R.; and Fan, G. 2024.
\newblock Sequential asynchronous action coordination in multi-agent systems: A stackelberg decision transformer approach.
\newblock In \emph{Forty-first International Conference on Machine Learning}.

\bibitem[{Zhao et~al.(2023)Zhao, Zhao, Li, Kannala, and Pajarinen}]{zhao2023optimistic}
Zhao, W.; Zhao, Y.; Li, Z.; Kannala, J.; and Pajarinen, J. 2023.
\newblock Optimistic multi-agent policy gradient.
\newblock \emph{arXiv preprint arXiv:2311.01953}.

\bibitem[{Zhao et~al.(2025)Zhao, Zhang, Chen, Zhang, Wang, and Zhou}]{zhao2025sequence}
Zhao, Z.; Zhang, Y.; Chen, W.; Zhang, F.; Wang, S.; and Zhou, Y. 2025.
\newblock Sequence Value Decomposition Transformer for Cooperative Multi-Agent Reinforcement Learning.
\newblock \emph{Information Sciences}, 122514.

\bibitem[{Zhong et~al.(2024)Zhong, Kuba, Feng, Hu, Ji, and Yang}]{zhong2024heterogeneous}
Zhong, Y.; Kuba, J.~G.; Feng, X.; Hu, S.; Ji, J.; and Yang, Y. 2024.
\newblock Heterogeneous-agent reinforcement learning.
\newblock \emph{Journal of Machine Learning Research}, 25(32): 1--67.

\end{thebibliography}

\clearpage

\end{document}